\documentclass[journal=jacsat,manuscript=article,layout=twocolumn]{achemso}

\usepackage{chemformula} 
\usepackage[T1]{fontenc} 
\usepackage{adjustbox}
\usepackage{graphicx}
\usepackage{color,soul}

\sethlcolor{white}

\author{James Houston}
\author{Frank G. Glavin}
\author{Michael G. Madden}
\email{frank.glavin@nuigalway.ie, michael.madden@nuigalway.ie}

\affiliation[National University of Ireland, Galway]
{School of Computer Science,\\ National University of Ireland, Galway.\\ H91 TK33}

\title{Robust Classification of High-Dimensional Spectroscopy Data Using Deep Learning \\ and Data Synthesis}


\begin{document}

%





\begin{abstract}
 This paper presents a new approach to classification of high dimensional spectroscopy data and demonstrates that it outperforms other current state-of-the art approaches. The specific task we consider is identifying whether samples contain chlorinated solvents or not, based on their Raman spectra. We also examine robustness to classification of outlier samples that are not represented in the training set (negative outliers). A novel application of a locally-connected neural network (NN) for the binary classification of spectroscopy data is proposed and demonstrated to yield improved accuracy over traditionally popular algorithms. Additionally, we present the ability to further increase the accuracy of the locally-connected NN algorithm through the use of synthetic training spectra and we investigate the use of autoencoder based one-class classifiers and outlier detectors. Finally, a two-step classification process is presented as an alternative to the binary and one-class classification paradigms. This process combines the locally-connected NN classifier, the use of synthetic training data, and an autoencoder based outlier detector to produce a model which is shown to both produce high classification accuracy, and be robust to the presence of negative outliers.
\end{abstract}

\section{Introduction}
%
%
%
%
The work presented in this paper concerns the task of classifying chemical samples by way of their Raman spectra. Specifically, samples are classified as to whether they contain chlorinated solvents (positive case), or do not contain chlorinated solvents (negative case). This is achieved through the use of machine learning algorithms, whereby a classifier is learned from a set of correctly-labelled training instances.  Since the negative case encompasses all samples other than chlorinated solvents, it is not feasible to have a negative training set that comprehensively represents the full range of possible negative instances. For this reason, two desirable properties of classification models are addressed in this work:
\begin{enumerate}
\item High classification accuracy on positive and negative test instances which are similar to those in the training set.
\item Robust performance when presented with negative outlier instances that are not well represented by the negative instances in the training set.
\end{enumerate}

\section{Background Information}
\label{sec:background}
\subsection{Analysis of Spectroscopic Data}
\label{sec:analysisSpec}
The analysis of spectroscopic data to obtain either quantitative or qualitative insight is a common task in the natural sciences. Raman spectroscopy is one particular spectroscopic technique used to observe low-frequency modes in a molecular system, such as rotational or vibrational modes \cite{cite1}. It relies on measuring the inelastic scattering of an intense beam of monochromatic light. Inelastic scattering, or Raman scattering, occurs when the scattered photon is shifted in energy compared with the incident photon, due to interaction of the photon with a vibrational or rotational mode of the illuminated molecule. By measuring the energy shifts at which light is most intensely Raman scattered, the presence of modes at particular energies can be determined. As such, the pattern of Raman intensities at different energies can be used as a `\emph{fingerprint}' to determine the presence of certain molecules. The shifts in energies are typically reported in terms of the change in wave number of the scattered photons. The wave number is the spatial frequency of a wave, measured in cycles per unit distance. It can be directly related to the energy of a photon. 
\subsection{Binary Classification}
\label{sec:binClass}
In machine learning, multi-class classification is the task of accurately assigning unseen examples with the correct class label from a predefined set of class labels. A classifier is constructed through inductive learning, which entails observations of correctly-labelled training data, each of which is a a set of values for a fixed set of features or attributes. The exact nature of this learning process is determined by the particular learning algorithm used. A simple example would be predicting if a student will pass a particular exam based on their attendance record and average exam score across other courses. Binary classification is a specifi case of multi-class classification in which there are only two classes, commonly referred to as the positive class and negative class.
\subsection{Classification of Raman Spectra}
\label{sec:specClass}
The primary task considered in this work is the identification of the presence of certain target compounds in a sample through analysis of its Raman spectrum. The set of target compounds comprises three chlorinated solvents. Differentiation of chlorinated waste from non-chlorinated waste is of significant practical importance due to different requirements of disposal of chlorinated and non-chlorinated laboratory waste \cite{cite2}. \\
\indent In a machine learning context this can be framed as a binary classification task, with the positive class corresponding to samples that contain the compounds of interest, and the negative class corresponding to samples that do not. From a  machine learning perspective, the dataset used in this work is of interest as an example of a classification task in a high-dimensional, continuous-value, low-data domain.  
\subsection{One-Class Classification}
\label{sec:oneClass}
In the context of this work, the negative class refers to any sample not containing a small set of target compounds. As such, it is much more poorly defined than the positive class. In such a scenario it is difficult to have a training set of negative examples which adequately represents the enormous range of possible negative samples. As a result, the robustness of binary classification methods to the introduction of unexpected negative outliers to the test set has been tested and compared to that of a one-class classifier (OCC) \cite{cite3}. A one-class classifier is an alternative to a conventional binary or multi-class classifier. They are also referred to as one-sided classifiers or single-class classifiers. In OCC algorithms the model is trained on a single, well defined class, called the positive or target class. The model learns to differentiate members of this class from all other instances. The formulation of a task into a one-class classification problem lends inherent robustness to the presence of negative outliers, since all negative instances are define as being outliers of the positive class. It has been noted that, in general, the problem of one-class classification is more difficult than binary or multi-class classification as the decision boundary is only supported from one side \cite{cite50}. Anomaly/Novelty/Outlier detection and PU (positive and unlabeled) learning are closely related to one-class classification in that they deal with a single, well-defined target dataset. 
\subsection{Artificial Neural Networks and Deep Learning}
\label{sec:ANN_DL}
Artificial neural networks are a type of learning algorithm loosely inspired by biological neural networks in the brain \cite{cite4}. A basic feed-forward neural network comprises an acyclic graph of nodes organized in layers. The first and last layers are the inputs and outputs, respectively. Any additional layers between these are referred to as hidden layers. In a conventional architecture every node in a layer will be connected to every node in the previous layer. The z-value at a particular node is the sum of the activations of the nodes in the previous layer times the weights of the connections between the nodes, plus a bias value. The activation value of the node is then determined by passing the z-value through an activation function, such as a \emph{sigmoid} function. The activations of the input layer are simply the input instance feature values. Further discussion of the technical details can be found in Russell and Norvig \cite{cite4}. By adjusting the weights and biases, a neural network may be trained to produce a certain output, such as zero or one in a binary classification use-case, for a certain input, such as negative or positive instances. A set of weights and biases to achieve this may be obtained by training on labelled data using the backpropagation algorithm, whereby error in the output is back-propagated through the network and the weights adjusted according to their contribution this error. \\
\indent Deep learning refers to a broad family of machine learning techniques, typically employing neural network architectures with multiple hidden layers \cite{cite5}. Deep learning architectures have had particularly successful application in perception tasks such as audio recognition, speech recognition and computer vision as well as difficult pattern recognition tasks such as drug design, bioinformatics and machine translation \cite{cite6}. \\  
\indent An important aspect in the effectiveness of deep learning is the integrated approach to feature engineering. Rather than manually designing features to present to the learning algorithm, the first layers in a deeper architecture learn appropriate features in order to perform the desired task. Many forms of neural network architecture are used within the field of deep learning, with appropriate architecture design for the type of task being vitally important to the ability of the algorithm to learn effectively. For example, Convolutional Neural Networks are often used in image classification \cite{cite51}, while Recurrent Neural Networks find applications in automatic speech recognition \cite{cite52}.  
\subsection{Data Synthesis}
\label{sec:dataSynt}
Data synthesis, or data augmentation, is an established technique in machine learning for the generation of additional training data from an existing dataset. It is often used to improve model robustness by simulating variance not well represented in the available training data. It is also commonly used in improving classifier performance by better sampling the feature space of each class of data available, thus enforcing greater generalization and improved model performance. \\
\indent Pattern recognition tasks in data sparse domains can prove difficult as there may not be sufficient variance in the available data for a model to learn good generalization. For a classification task, data augmentation in the form of label-preserving transformations may be applied to the existing training data in order to create more variation within training classes. 
\section{Literature Review}
\label{sec:litRev}
\subsection{Applications of Machine Learning to Spectroscopy Data}
\label{sec:appMLtoSpec}
Madden and Ryder \cite{cite7} explore the use of k-Nearest Neighbours and Feed-Forward Neural Network regression models in the identification and quantification of illicit materials based on Raman spectra. They note the difficulty in applying predictive models to high-dimensional data where the number of attributes per sample may greatly exceed the number of samples. They describe an approach to dimension reduction utilizing genetic algorithms to determine an appropriate subset of attributes. Additionally, the use of ensemble methods to achieve greater accuracy in prediction is demonstrated. \\
\indent O'Connell \emph{et al.} \cite{cite8} propose the use of Principal Component Analysis (PCA) and support vector machines (SVMs) to determine the presence of a target analyte, acetaminophine, from the Raman spectrum of a solid mixture. A range of preprocessing techniques, including first derivative and normalization transformations are employed. Following pre-processing, presence of the target analyte is discriminated using PCA, Prinicpal Component Regression (PCR) and SVMs. Superior performance of SVMs is noted, particularly when raw data, without the preprocessing steps, is used. \\
\indent Glavin and Madden \cite{cite3} explore the effect of unexpected negative outliers in the test data on the classification of Raman Spectra. A One-Sided kNN algorithm is compared to two conventional binary classification algorithms, an SVM and a binary kNN. Two scenarios are considered: `Expected' test data only, and `Unexpected' and `Expected' test data. The classification task considered is the identification of the presence of chlorinated solvents in a solvent mixture. Greater robustness to `unexpected' test instances is shown in the one-class classifier, although better classification accuracy is displayed by the two binary classifiers in the absence of `unexpected' test instances. \\
\indent Conroy \emph{et al.} \cite{cite2} provide a direct comparison of chemometric and machine learning techniques in qualifying and quantifying the presence of chlorinated solvents in a sample based on Raman spectra. The machine learning techniques were found to produce better classification results with the Ripper rule-learning algorithm, SVM, and C4.5 producing the best results. The SVM algorithm is found to give the best performance in detecting if any chlorinated solvent is present. \\ 
\indent Howley \cite{cite9} demonstrates the use of Principal Component Analysis (PCA) with machine learning algorithms including SVMs in quantifying materials from their Raman spectra. The author reports improved results using this method over conventional chemometric techniques such as Principal Component Regression (PCR). \\
\indent Zhao \emph{et al.} \cite{cite10} compare two machine learning approaches to categorizing tea from their infra-red spectra. They use an artificial neural network as well as an SVM with PCA to reduce the dimensionality of the spectral data. The author notes superior performance of the SVM, attributing this to the better generalization in comparison to the neural network approach. \\
\indent Sheinfeld \emph{et al.} \cite{cite11} use a set of pre-trained neural networks to determine the levels of radioactivity in building materials from their gamma spectra. The authors report that best results are achieved using a fully connected architecture comprising 3 hidden layers and 1,144 total weight elements. Additionally, an ensemble of individually trained neural networks is employed with a tailored averaging algorithm used to produce the final results.  The authors report parity with the results of traditional methods. It should be noted that in this work significant pre-processing of the raw spectra is carried out, with feature reduction being achieved by focusing on 19 selected regions of interest. \\
\indent Chen \emph{et al.} \cite{cite12} use a hybrid framework of principal component analysis, a stacked autoencoder deep learning architecture, and a logistic regression model in classification of hyper-spectral data. They report competitive performance on a widely used dataset. The hyper-spectral data contains both spatial and spectral continuous data. They extract the spatial and spectral data separately and feed the resultant vectors into a single stacked autoencoder network. The output vector of the neural network is then passed to a logistic regression model and a classification is produced. A fully connected architecture is employed in the stacked autoencoder. \\
\indent The application of fit-to-purpose deep neural network architectures to spectroscopic classification problems has received less attention relative to domains such as image classification or speech recognition.

\subsection{Data Synthesis}
\label{sec:dataSynthesis}
\begin{figure}[!t]
\centering
\includegraphics[width=3.3in]{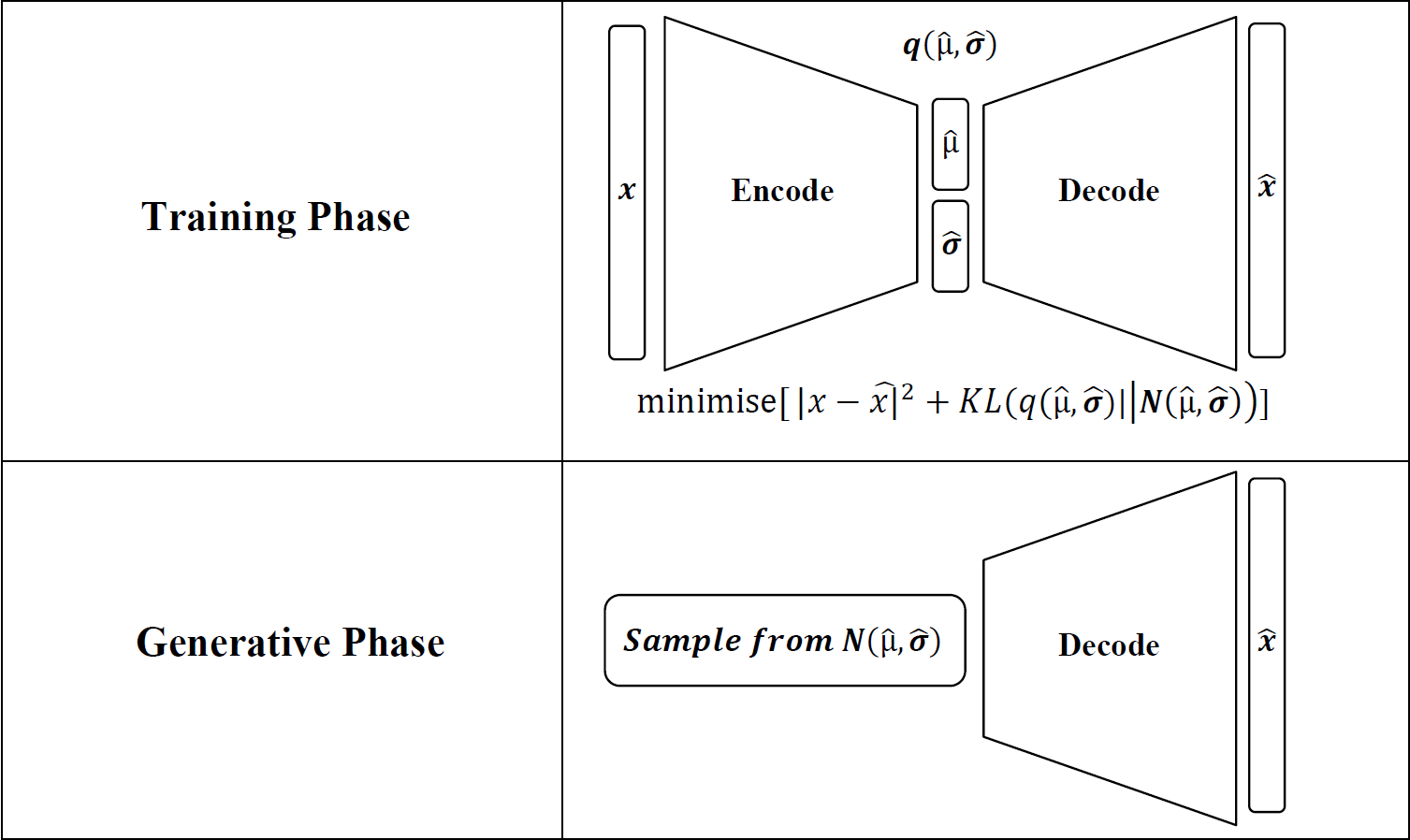}
\caption{Architecture of Variational Autoencoder in both the training and generative phases.}
\label{fig1}
\end{figure}

Data synthesis, or data augmentation, techniques have been applied extensively in the areas of deep learning and perception tasks due to the requirement of large volumes of training data which, in many cases, may not exist or be available. In the image recognition and Computer Vision domain, transformations such as translation, reflection, and deformation are used to generate augmented training images, with a demonstrated improvement in model performance \cite{cite6}\cite{cite13}\cite{cite14}. In the speech recognition domain, data augmentation has primarily been used to produce more robust models by simulating variance and nuisance factors in the training data \cite{cite15}\cite{cite16}\cite{cite17}. Cui \emph{et al.} \cite{cite18} use label preserving transformations to deal with data sparsity for deep neural network and convolutional neural network based acoustic modelling. \\
\indent Data augmentation has also seen application to spectroscopic data. Georgouli \emph{et al.} \cite{cite19} describe the application of a four block data augmentation process to spectroscopic data of vegetable oils for training in a classification task. The independent augmentation procedures comprised a spectrum blender (creating artificial mixtures from a weighted sum of input samples), a spectral intensifier, shifting on the x-axis, and the addition of Gaussian white noise on the y-axis. The authors test the effect of data augmentation on the performance of statistical methods for identifying vegetable oil admixtures. They report improved classifier performance and robustness through the use of data augmentation.
S{\'a}iz-Abajo \emph{et al.} \cite{cite20} implement a variety of noise simulation methods in the application of analysing spectroscopic data. These methods include additive noise (adding zero mean Gaussian noise to each point on spectrum), intensity dependent noise (standard deviation of Gaussian noise is dependent on the spectral intensity), and multiplicative noise (spectra is multiplied with log normal noise of mean 1). \\
\indent A non-domain specific method for generating synthesised new data from an existing dataset is achieved through the use of a variational autoencoder (VAE). Developed in 2014 by Kingma and Welling \cite{cite21}, VAEs are a form of generative autoencoder in the sense that they can generate new instances which appear to come from the same sample distribution as the training set. As with any autoencoder they work by learning to reproduce the training data and, in the process, produce an encoding of the training data in the form of the hidden layer activations. However unlike standard autoencoders which produce a direct coding for each instance, VAEs instead produce a coding distribution. A term is added to the cost function which measures the Kullback-Leibler divergence between the produced coding distribution and the desired distribution (eg. mean zero Gaussian distribution with variance equal to one). Instead of direct codings, distribution parameter codings $\hat{\mu}$ and $\hat{\sigma}$ are produced. In this way, the VAE learns to code and decode from a latent space in which the training samples are normally distributed. The decoding portion of the trained VAE can then be used as a generative model by simply sampling within the latent space according to the learned $\hat{\mu}$ and $\hat{\sigma}$ parameters and decoding to produce a new  $\hat{x}$. Figure \ref{fig1} shows a diagrammatic representation of a VAE in both the training and generative phases.\\
\indent Nishizaki \cite{cite22} describes the use of a VAE for the purpose of data augmentation in the speech recognition domain. The authors investigate the effectiveness of data augmentation using a VAE framework and show that latent variable-based features can be utilized in automatic speech recognition.

\subsection{One-Class Classification Algorithms and Applications}
\label{sec:occ}
Khan and Madden \cite{cite23} propose a broad categorisation of OCC algorithms as either One-Class Support Vector Machine (OSVM) based or non-OSVM based. They argue that the levels of advancement and variation within OSVM-based algorithms justify the labelling of OSVMs as a subfield in their own right. \\
\indent Tax and Duin \cite{cite24}\cite{cite25} introduce the method of Support Vector Data Description (SVDD). A hyper-sphere of minimum radius is constructed such that it encompasses all, or nearly all, of the positive training cases. As with conventional SVMs kernel functions may be used to transform the space in which a boundary is found. It is noted that the SVDD method becomes inefficient with high-dimensional data and does not work well in cases where there is a large variation in the density of the target class instances. \\
\indent Tax and Duin \cite{cite26} propose the use of artificially generated outliers in the optimization of OSVM parameters. They generate example outliers uniformly in a hypersphere in and around the target class. They determine that this process is feasible for up to 30 dimensions.\\ 
\indent Sch{\"o}lkopf \emph{et al.} \cite{cite27} propose an alternative OSVM technique in which a separating hyper-plane is constructed as opposed to a hyper-sphere. The hyper-plane is constructed such that the distance from the origin is maximised and it can separate the data containing region from the region containing no data. \\
\indent A variety of non-OSVM one-class classification algorithms have been developed, with many being variations of existing multi-class classification techniques. As reported by Khan and Madden \cite{cite23}, De Comité \emph{et al.} \cite{cite28} modify a conventional C4.5 decision tree algorithm to train using only positive and unlabelled data. Letouzey \emph{et al.} \cite{cite29} similarly propose a decision tree which learns from positive and unlabelled data, here called POSC4.5. Both algorithms are demonstrated to give competitive performance on UCI datasets. It has been noted that such rule base methods may not be appropriate for use with continuous features and high dimensional data \cite{cite30}.\\
\indent Tax \cite{cite31} demonstrates a one-class Nearest Neighbour algorithm which he calls Nearest Neighbour description (NN-d). The algorithm works by calculating the distance from a test instance, $z$, to its nearest training instance, $NNtr(z)$. This distance is then compared to the distance from $NNtr(z)$ to its nearest training instance, $NNtr(NNtr(z))$. The ratio of the two distances is used in determining if a test instance belongs to the positive class. The threshold used, as well as the number of nearest neighbours, may be varied.\\
\indent Khan \cite{cite32} proposes the use of two one-class kNN variants in the identification of chlorinated solvents. One method finds a single neighbourhood for $k$ nearest neighbours while the other method finds multiple localised neighbourhoods for each of $k$ neighbours. Both methods employ a variety of kernels as a distance metric in place of Euclidean distance. The kernels used include polynomial (degree 1 and 2), radial basis function, Spectral Linear Kernel \cite{cite9} and Weighted Spectral Linear Kernel \cite{cite33}. They report improved performance over standard NN-d methods. \\
\indent Neural network based OCC algorithms have also been developed. Skabar \cite{cite34} presents a description of a feed-forward neural network which learns from a corpus of positive and unlabeled data. The application of this technique in predicting the location of mineral deposits demonstrates the ability of the neural network to create a decision boundary between positive and negative classes without any explicitly labelled negative training instances. \\
\indent In recent years, autoencoders have become a popular method for one-class classification and outlier detection \cite{cite35}. Manevitz and Yousef \cite{cite36}\cite{cite37} employ an autoencoder network with one hidden layer and a bottleneck, such that the number of nodes in the hidden layer is greatly reduced compared to the input / output layers. The bottleneck prevents the model from learning the full identity function for the input space. The model is forced to approximate the identity function only for input vectors is the sub-space of vectors which are similar to the training set. In this way, the ability of the trained model to reconstruct an input instance can be used to differentiate positive class (target) instances from negative class (outlier) instances.\\
\indent Hawkins \emph{et al.} \cite{cite38} demonstrate the effectiveness of using replicator neural networks for outlier detection on two publically available data records. A step-wise activation function is used in the hidden layer to divide the continuous data into clusters.\\
\indent Marchi \emph{et al.} \cite{cite39} propose the use of a denoising autoencoder as part of a novel approach for automatic acoustic novelty detection. They use the reconstruction error of a denoising autoencoder to detect anomalies (outliers). The denoising autoencoder is trained to reconstruct the input from a corrupted (noisy) version of the input, with the idea being this forces the hidden layer to learn more robust features. They train the model using normal household audio and test its ability to detect outliers (screams, alarms etc.) unseen in the training phase. They report improved results over state of the art.\\
\indent Sakurada and Yairi \cite{cite40} provide a comparison of anomaly detectors including autoencoders, denoising autoencoders and linear PCA. The models were tested on artificial data. The potential to increase the accuracy of autoencoders by extending them to denoising autoencoders is noted. Additionally, it is demonstrated that autoencoders are able to detect subtle anomalies which PCA fails to detect.\\
\indent Erfani \emph{et al.} \cite{cite41} present a combined deep-learning and one-class SVM method for anomaly detection. A deep belief network is employed to engineer robust, low dimensional features. These features are then used as input to an OSVM algorithm. The algorithm is tested in a variety of high-dimensional problem domains. The authors report performance comparable with a deep autoencoder.
Cao \emph{et al.} \cite{cite42} describe a hybrid autoencoder and density estimation model for anomaly detection. A compression autoencoder is trained on the positive data. However, rather than using the reconstruction error to detect outliers, a density estimation model is used to determine the activation of the hidden layer. Low density of the hidden layer is considered to be indicative of outliers. The authors report improvement on best known results on 3 of the 4 datasets tested. \\
\indent While the applicability of kNN and OSVM based one-class classification approaches to classification tasks on spectroscopic data has been explored \cite{cite3}\cite{cite32}\cite{cite43}, the use of neural network based approaches such as autoencoders and denoising autoencoders on spectroscopic data is not as well understood. 

\section{Dataset and Classification Algorithms}
\subsection{Dataset}
\label{sec:dataset}
The principal dataset used in this study comprises 230 Raman spectra of solvents and solvent mixtures. The dataset was originally compiled for research conducted by Conroy \emph{et al.} \cite{cite2}. The 230 samples in the dataset were created from a set of 25 solvents. It includes pure samples as well as a non-exhaustive selection of 2, 3, 4, and 5 solvent mixtures as shown in Table \ref{table1}. All spectra in the dataset are labelled as to whether the sample contains a chlorinated solvent or not. Chlorinated spectra will be referred to as positive instances and non-chlorinated spectra referred to as negative instances in this work.

\begin{table}[h!]
\begin{adjustbox}{max width=0.46\textwidth}
\begin{tabular}{*{4}{|c}|}
   \hline
   & Chlorinated & Non-chlorinated & Total \\ 
   \hline
\hline
	Pure Solvents & 6 & 24 & 30 \\ \hline
	Two Solvent Mixtures & 96 & 23 & 119 \\ \hline
	Three Solvent Mixtures & 40 & 12 & 52 \\ \hline
	Four Solvent Mixtures & 12 & 10 & 22 \\ \hline
	Five Solvent Mixtures & 0 & 7 & 7 \\ \hline
	Total & 154 & 76 & 230 \\ \hline
 \end{tabular}
 \end{adjustbox}
\caption{Breakdown of sample compositions (source: Glavin and Madden \cite{cite3})}
\label{table1}
\end{table}

\begin{figure}[!t]
\centering
\includegraphics[width=3.3in]{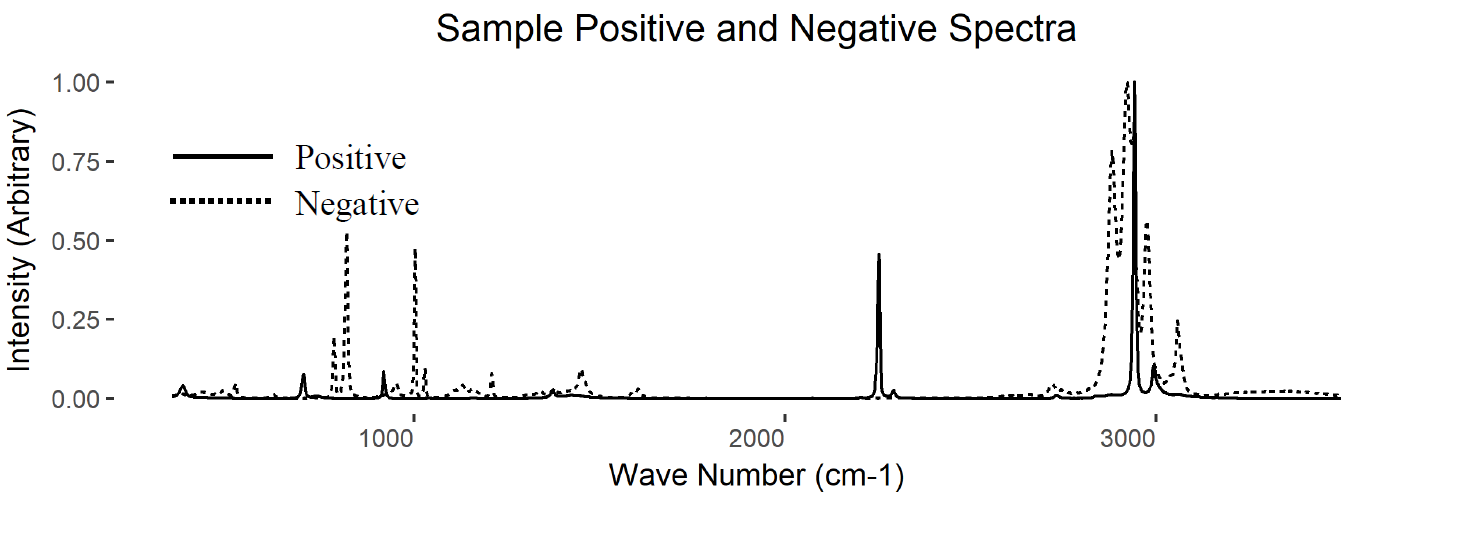}
\caption{Example of positive and negative spectra, (chlorinated and non-chlorinated).}
\label{fig2}
\end{figure}

The spectra themselves comprise a set of Raman scattering intensity readings at 2,473 wave number values, measured in cm-1. In a machine learning context, each spectrum is treated as an instance with 2,473 features. Each spectrum has been normalized such that the minimum and maximum intensity values are equal to 0 and 1 respectively. Figure \ref{fig2} shows an example of both a positive and a negative spectrum taken from the principal dataset.

\subsection{Classification Algorithms}
\label{sec:classAlgs}
A number of binary classification algorithms are applied in order to assess their suitability in differentiating positive and negative spectra. The algorithms applied are a kNN classifier, SVM classifier, Decision Tree classifier, Guassian Na\"{\i}ve Bayes, Fully Connected Neural Network, and a Locally Connected Neural Network. These algorithms are compared on the basis of classification accuracy attained on an unseen test set by the trained model. The aim is to determine a baseline of classification performance obtainable on the raw data using an array of algorithms. The performance of the Fully Connected Neural Network and Locally Connected Neural Network algorithms can then be assessed through comparison. A brief description of each algorithm is given below. 
\subsubsection{k-Nearest Neighbours (kNN) classifier}
The kNN algorithm is one of the simplest examples of a learning algorithm. The training and testing instances are represented as points in $n$-dimensional space, where n corresponds to the number of features in the data. Rather than learning an explicit prediction model from the data, the training instances are simply stored. When presented with a test instance, the distance to each stored training instance is calculated. Common distance metrics include \emph{Euclidean} distance and \emph{Manhattan} distance. The $k$ nearest neighbours are considered, where $k$ is a positive integer, and can be tuning parameter for the specific application. In the case of a binary classification task,a majority vote of the class labels of the $k$ nearest neighbours is used to assign a class prediction to the test case. 
\subsubsection{Support Vector Machine (SVM) classifier}
\label{sec:SVM}
A support vector machine operates under the principle of finding the maximum margin separating hyper-plane. That is, the hyper-plane in $n$-dimensional space, separating the positive and negative instances, such that the nearest instances on each side of the boundary are as far as possible from the boundary. This principle of finding the widest separating boundary promotes maximum generalization of the model. In most practical scenarios, a perfectly separating hyper-plane does not exist. In such cases, the constraint of perfectly separating the data is relaxed with the introduction of slack variables, corresponding to the existence of training instances within the margins of the separating hyperplane. The sum of these slack variables is added to the optimization cost function with a weighting parameter $C$, which parameter may be tuned in order to balance generalization of the model with good performance on the training set. \\
\indent In cases where a good separating hyper-plane does not exist in the original attribute space then the problem may be transformed to a higher-dimensional space involving non-linear combinations of the original attributes. This process is performed efficiently in SVMs through the use of the kernel trick, which avoids the need to explicitly transform all training cases into the higher dimensional representation. Commonly-used kernels include the Radial Basis Function (RBF), polynomial function (of varying degrees) and the linear kernel, which corresponds to the original feature space of the data. 
\subsubsection{Decision Tree classifier}
\label{sec:decisionTree}
A decision tree classifier operates by learning a tree of rules with which to classify an instance. Each branching point in the tree is a conditional rule involving an attribute and a threshold. The branches eventually end in leaves which have associated classifications. The nature and order of the rules are determined from the training data according to a greedy algorithm. At each branch, starting at the root, the attribute and associated threshold value which best separate the training instances by class is chosen. The data is then split according to the this rule and the same process is applied recursively to each subsets of data from the split. Since the data is split greedily, the final tree is not guaranteed to be optimal. A scoring function is used to determine which candidate split is the `best' one; the two most common scoring functions are Information Gain and Gini Index. In order to prevent overfitting, a maximum tree depth is often employed. By varying the maximum tree depth the balance between generalization and performance on the training set can be tuned.
\subsubsection{Gaussian Na\"{\i}ve Bayes classifier}
\label{sec:gasNaive}
A Na\"{\i}ve Bayes classifier operates by estimating the probability that a test instance belongs to a certain class. In order to achieve this, it is necessary to express this probability as a function of the feature values of the instance. An assumption is made that all feature are independent and Bayes' theorem is applied to relate the probability of a class given a set of attributes to the probability of a set of attributes given a class multiplied by the prior probability of that class. The prior probability of a class is easily estimated from the ratios present in the training set. Additionally, an estimation of the probability density function for each continuous feature for each class is estimated from the data. In a case where Guassian kernel is used for probability density , the classifier is referred to as a Gaussian Na\"{\i}ve Bayes classifier.
\subsubsection{Fully Connected Neural Network (FCNN) classifier}
\label{sec:fcnn}
In a binary classification task, a single output node is used. In training, the network is trained to output 1 for positive instances and 0 for negative instances. In prediction, the output node produces a numeric value, to which a threshold is applied (usually 0.5) in order to determine the predicted class. Training is typically carried out for many epochs, with an epoch meaning the network has been trained once on every training instance. 
\subsubsection{Locally Connected Neural Network (LCNN) classifier}
\label{sec:lcnn}
In a locally connected layer, each node is only connected to a subset of nodes in the previous layer, known as its receptive field. Locally connected layers are typically employed where there is correlation between inputs related to their ordering, such as adjacent pixels in an image or, in the case of spectra, intensity readings at consecutive wave numbers. Having nodes only connect to a receptive field of nodes in the previous layer can greatly reduce the number of parameters to be trained. The appropriateness of a locally connected architecture on raw spectroscopic data is argued here for the following reasons:
\begin{enumerate} 
\item	Conventional chemometric and machine learning techniques may ignore subtle patterns present in the data resulting in suboptimal classification performance. This is particularly true when dimensionality reduction techniques such as PCA or the selection of regions of interest are used.  
	\item Due to the high dimensionality of the data, the use of a fully connected layer results in a very large number of training parameters even when the dimensionality of the first hidden layer is small.
\end{enumerate}
\hl{We note that other authors have reported success in classification of Raman spectra using deep networks that include convolutional and pooling layers} \cite{cite55, cite56}. \hl{In our work, we have chosen to use a simpler architecture with locally connected layers rather than a full convolutional neural network.}

\subsection{Training, Testing and Validation}
\label{sec:trainTestVal}
A random, stratified split of the data is used to produce an experimental set and a validation set in a 75:25 ratio. The experimental set is further split into 5 cross-validation folds, again stratified. The cross-validation folds are used in all experiments in this paper. The validation set is used to test the final model performance discussed in the Two-Step Classification Section. Table \ref{table2} shows the breakdown on positive and negative instances in each fold.

\begin{table}[h!]
\begin{adjustbox}{max width=0.46\textwidth}
\begin{tabular}{|c|c|c|c|c|}
 \hline
	Data Split & \multicolumn{2}{c}{Training} &   \multicolumn{2}{c}{Testing}  \\ \hline
	 & Positive  & Negative & Positive & Negative \\ \hline
	Fold 1 & 92 & 45 & 23 & 12 \\ \hline
	Fold 2 & 92 & 45 & 23 & 12 \\ \hline
	Fold 3 & 92 & 45 & 23 & 11 \\ \hline
	Fold 4 & 92 & 46 & 23 & 11 \\ \hline
	Fold 5 & 92 & 46 & 23 & 11 \\ \hline
	Validation & N/A & N/A & 39 & 19 \\ \hline
 \end{tabular}
 \end{adjustbox}
\caption{Class breakdown of the validation set and 5-fold cross-validation splits}
\label{table2}
\end{table}

\subsubsection{Model Training and Testing}
\label{sec:modTrainTest}
\begin{figure}[!h]
\centering
\includegraphics[width=3.3in]{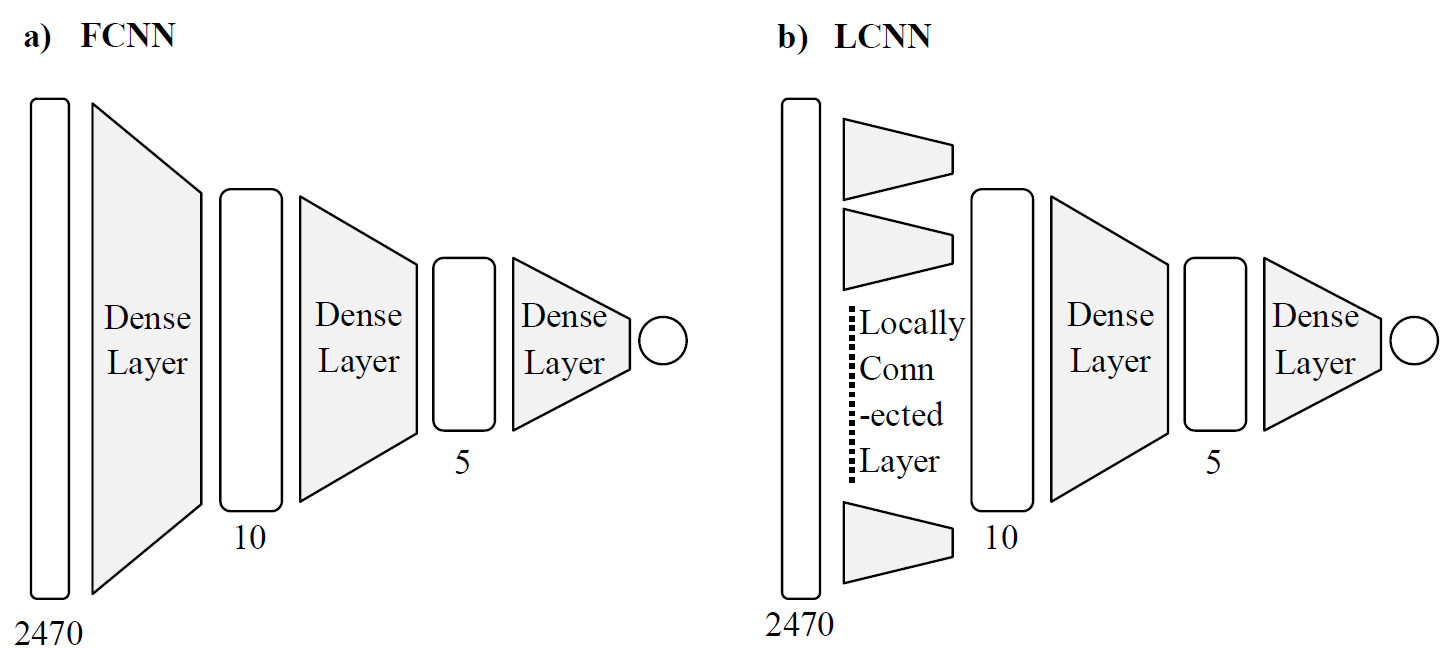}
\caption{Architecture of a) the FCNN and b) the LCNN used.}
\label{fig3}
\end{figure}

All models are trained on each of the 5 training folds and tested on the respective test folds. The mean classification accuracies and associated standard error of the mean across all 5 folds are reported in this work.\\ 
\indent Scikit-Learn \cite{cite44} implementations of kNN, SVM, Decision Tree, and Guassian Na\"{\i}ve Bayes Classifier algorithms are used. Hyper-parameter tuning of these algorithms is performed via a grid search, with the parameters yielding the highest mean accuracy on the test sets being selected. The final classification algorithms tested are a fully connected neural network (FCNN) and a locally connected neural network (LCNN). Both algorithms are implemented using the Keras package \cite{cite45}. The architectures of both models are described below: \\
\textbf{FCNN:}
A fully connected layer is one in which every node in a layer is connected to every node in the previous layer. They also commonly referred to as dense layers. The FCNN architecture used comprises three dense layers, the first connecting the input nodes to the first hidden layer of 10 nodes, then to the second hidden layer of 5 nodes, and finally to the single output node. In the first hidden layer, $tanh$ activation was used, in the second hidden layer, $ReLU$ activation was used, and for the output layer, $sigmoid$ activation was used. \hl{We apply $L1$ regularization with $\lambda=10^{-5}$ in the first hidden layer only.} The loss is calculated using \hl{binary cross-entropy}, and Adam optimization \cite{cite53} is used. The models are trained with a batch size of 5 for 100,000 epochs. \\
\textbf{LCNN:}
The LCNN architecture and training process are identical to the FCNN, save for the use of a locally connected layer going from the inputs to the first hidden layer. All layer activations and training parameters are kept consistent with \hl{those of the FCNN architecture, including $tanh$ activation and $L1$ regularization with $\lambda=10^{-5}$ in the first hidden layer.} 
Figure \ref{fig3} shows a diagrammatic representation of the FCNN and LCNN architectures used. 
The locally connected layer uses non-overlapping receptive fields of 247 nodes. Thus the input features are split into 10 receptive fields, each connecting to a single node in the first hidden layer.

The kNN, SVM, Decision Tree, and Gaussian Na\"{\i}ve Bayes algorithms used are deterministic, meaning given a particular training set they will arrive at the same model each time. In contrast, both the FCNN and the LCNN are non-deterministic due to the randomness of the initial weighting, the training process, and the absence of any guarantee of finding a global minimum. In practice it is possible to obtain reproducible results by setting the random number generator seed before generating and training these models. The models are generated and trained using 5 different seeds and the results averaged in order to better characterize their true behaviour. 
\subsection{Results}
\begin{figure}[!h]
\centering
\includegraphics[width=3.3in]{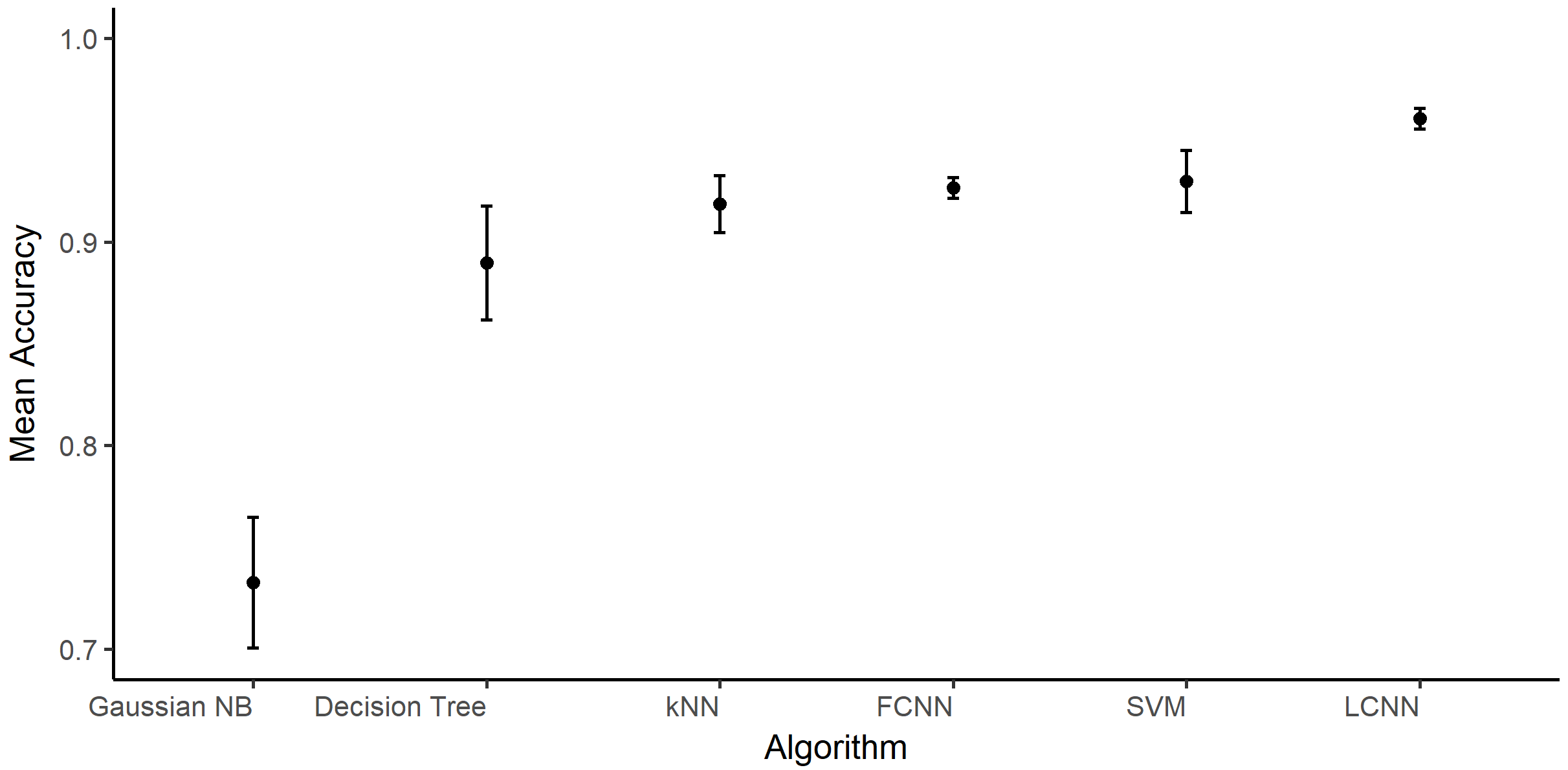}
\caption{Mean accuracy attained with variety of binary classification algorithms.}
\label{fig4}
\end{figure}

A summary of the results is shown in Table \ref{table3}. Figure \ref{fig4} shows the mean accuracy (and standard error of the mean) attained with each of the 6 classification algorithms that were tested. The accuracies shown correspond to the optimal hyper-parameters found via grid search (where applicable).

\begin{table}[h!]
\begin{adjustbox}{max width=0.46\textwidth}
\begin{tabular}{ | c | c | c | c | }
\hline
	Algorithm & Hyper-Parameters & Mean Accuracy & Std. Error \\ \hline
	Gaussian NB & N/A & 0.733 & 0.032 \\ \hline
	Decision Tree & Max Depth = 10 & 0.890 & 0.028 \\ \hline
	kNN & Manhatten Dist. $k$=1 & 0.919 & 0.014 \\ \hline
	SVM & linear kernel, $C$=10 & 0.930 & 0.015 \\ \hline
FCNN & N/A & \hl{0.930} & \hl{0.004} \\ \hline
LCNN & N/A & \hl{0.960} & \hl{0.004} \\ \hline
 \end{tabular}
\end{adjustbox}
\caption{Summary of binary classification algorithm performance}
\label{table3}
\end{table}

\begin{figure}[!h]
\centering
\includegraphics[width=3.3in]{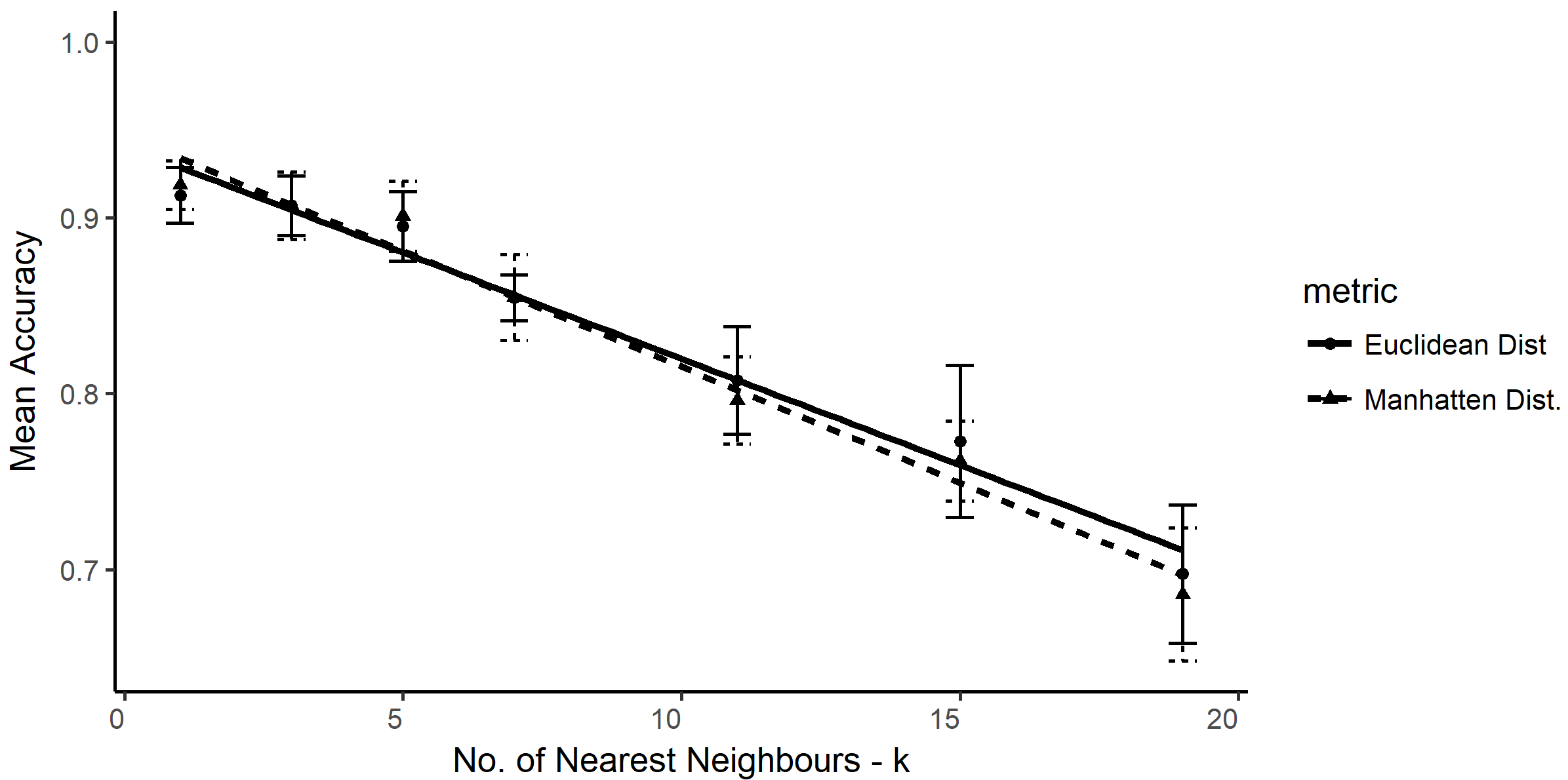}
\caption{Effect of distance metric and choice of $k$ on kNN classifier performance.}
\label{fig5}
\end{figure}

Gaussian Na\"{\i}ve Bayes shows the poorest performance by a large margin. Given the assumptions of Na\"{\i}ve Bayes (independence and relevance of all explanatory variables) it is not surprising that performance on a spectral dataset with high interdependence of variables is poor. The Decision Tree algorithm also yields relatively poor performance. A possible explanation of this could be the high degree of multi-collinearity in the dataset. In other words, multiple variables comprising the same peak will be highly linearly correlated with each other. A decision tree will tend to greedily use one of many collinear variables and discard the rest, potentially losing some subtle information in the data which other algorithms may exploit.\\    
\indent The kNN algorithm performs quite well considering the large ratio of explanatory variables to training instances. The results of the hyper-parameter grid search show very similar performance between distance metrics, with a slightly higher maximum accuracy attained with Manhattan distance over Euclidean distance. A much larger effect on performance is seen with the choice of $k$, the number of nearest neighbors considered. The best performance is achieved with a value of $k$=1 (See Figure \ref{fig5}). This result demonstrates that the majority of test instances are best characterized by their single nearest neighbour in the training set. 
 
\begin{figure}[!h]
\centering
\includegraphics[width=3.3in]{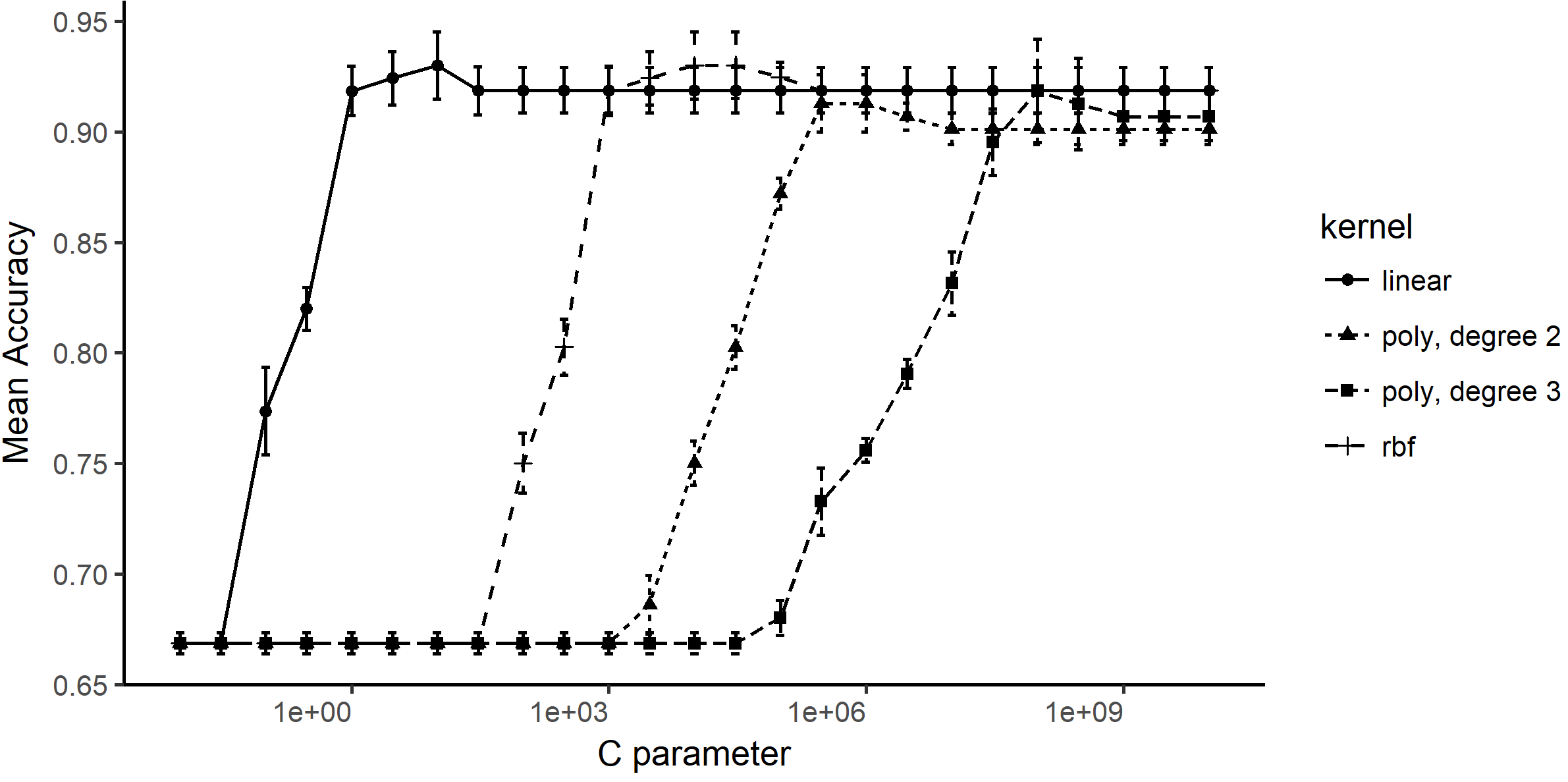}
\caption{Effect of C parameter and choice of kernel on SVM classifier performance.}
\label{fig6}
\end{figure}

The high accuracy achieved using an SVM classifier is expected given the suitability of the algorithm to high dimensional data, and is consistent with past research \cite{cite9, cite33}. Figure \ref{fig6} shows the results of the hyper-parameter grid search performed. An optimal range of $C$ parameter value can clearly be seen for each choice of kernel. The highest individual accuracies are attained using the linear kernel ($C=10$) and RBF kernel ($C$ = 10,000 or 30,000). Due to its simplicity and faster training time, the linear kernel SVM is used in all subsequent experiments. 
 
The FCNN yields very similar performance to the simpler SVM and kNN models, (\hl{mean accuracy 0.930$\pm$0.004}). From these results, no \hl{strong justification is found} 
for using a fully-connected neural network of this architecture on this data, taking into consideration the greatly increased training time incurred. The trained FCNN models display near 100\% percent accuracy on the training set, (mean accuracy 0.999$\pm$0.004). This result indicates that the algorithm is expressive enough to find the `true' boundary between positive and negative instances. However the disparity between performance on the training and testing sets indicates overfitting of the models to the training data.  The FCNN architecture used contains 24,771 trainable parameters, with 24,710 of those coming from the weights and biases of only the first layer. With only approximately 138 training instances for each model, the high ratio of parameters to training instances is a likely cause of this overfitting. The motivation for testing a locally connected architecture is in reducing the ratio of parameters to training instances. \\
The LCNN performs significantly better than any of the other models tested, (\hl{mean accuracy 0.960$\pm$0.004}). The use of the locally connected first layer reduces the total number of trainable parameters to 2,541. The trained models still display near 100\% percent accuracy on the training set, (mean accuracy 0.999$\pm$0.004). The large reduction in parameters appears to have the desired effect of increasing the generalization \hl{capability} of the model and reducing the severity of overfitting seen with the FCNN architecture. \\
\hl{While it might have been possible to further improve the performance of the FCNN and LCNN architectures by applying other adjustments (e.g. batch normalization, different activation functions or different regularization settings), our like-for-like comparison between the LCNN and FCNN architectures gives us confidence that both architectures would benefit similarly from such adjustments, and LCNN would still be the better of the two.}

\section{Data Synthesis and the Effect on Classification Accuracy}
\label{sec:dataSyntOnClass}
In this section, two methods of generating additional, synthetic training instances from the existing training sets are explored. The synthetic data are used for the training of the previously discussed kNN, SVM, and LCNN models. The potential for improving upon the observed classification accuracy of these algorithms, in particular the LCNN model, through the use of synthetic training data is tested. 
\subsection{Motivation for the use of Data Synthesis}
\label{sec:dataSyntMotiv}
The reduction in trainable parameters, and resulting improved classification accuracy, achieved through the use of a locally connected architecture, was discussed earlier. However, despite this reduction, the number of parameters still far exceeds the number of training instances and there is likely scope for improving classification accuracy further through the enforcement of better generalization.\\
\indent Further simplification of the learning model could be done to this end. However, at some point, simplification of the model may harm the learning algorithms ability to approximate the true class separation. Consider for example the comparative performance observed when using the SVM and the LCNN learners. An SVM will tend to generalize extremely well, however the LCNN performs notably better on the test data. This indicates that the performance of the SVM is being limited by its inability to express a more subtle decision boundary. \\
\indent Increasing the number and variety of training instances is a viable method for increasing generalization without sacrificing the expressiveness of the learning algorithm. By increasing the number of training instances the tendency to simply `memorise' the dataset is reduced. In deep learning applications, the use of very large training sets is common practice when learning to perform subtle perception or complicated pattern recognition tasks \cite{cite5}.\\
\indent In many practical cases the number of available, labelled, training instances is limited and the acquisition of further data may be impractical, time-consuming, or expensive. In such cases it may be desirable to generate artificial training instances from the limited available data. In this work, the total number of available Raman spectra is 230 and in most cases only $\approx$138 real spectra are used to train individual models. Thus, the use of artificial or synthetic training data is explored in an attempt to further improve classification accuracy of the LCNN model. For comparison, the effect of synthetic training data on the kNN and SVM models is also tested.
\subsubsection{Choice of Data Synthesis Methods}
\label{sec:choiceDS}
For synthetic data to be effective in improving the performance of a learning model, it needs to satisfy two criteria:
\begin{enumerate}
	\item The generated instances need to, in general, be somewhat distinct from any individual real training instances. If this is not the case then using the synthetic data is no different from simply duplicating instances in the training set. 
	\item The generated instances need to be sufficiently representative of the real instances, such that learning from them translates effectively to prediction on real test instances. 
	\end{enumerate}
In the case of a binary classification task, the data synthesis process must be label-preserving. That is, positive synthetic instances must be created from positive training data and negative synthetic instances must be created from negative training data in such a way that the characteristics of the classes are preserved. \\
\indent Two methods for generating synthetic data are considered. One is informed by some basic understanding of the data, the other is a more general, domain-independent method for data synthesis. They are briefly described below:\\
\textbf{Data Blending:}
Georgouli \emph{et al.} \cite{cite19} describe a spectrum blending technique for data augmentation on Fourier transform mid infrared (FTIR) spectroscopic data. The technique `blends' multiple parent spectra to create a new, synthetic spectrum. The blending process comprises calculating the weighted mean of the spectral intensities of the parent spectra at each point. The resultant blended spectrum can be considered to be in some way a representation of the Raman spectra which would resulting from a weighted mixture of the parent samples. This method is quite rigid in the types of variation it produces but does offer synthetic instances which are good representations of possible real data. \\
\textbf{Data Generation using a Variational Autoencoder (VAE):}
As discussed earlier, a VAE is a form of generative autoencoder in the sense that it can generate new instances which appear to come from the same sample distribution as the training set. By training separate VAEs on the positive and negative training data the resultant generative networks may be used as sources of label preserving synthetic data.
\subsection{Methodology}
\label{sec:methods}
\subsubsection{Spectrum Blender Data Generation}
\label{sec:specBlend}
\begin{figure}[!h]
\centering
\includegraphics[width=3.3in]{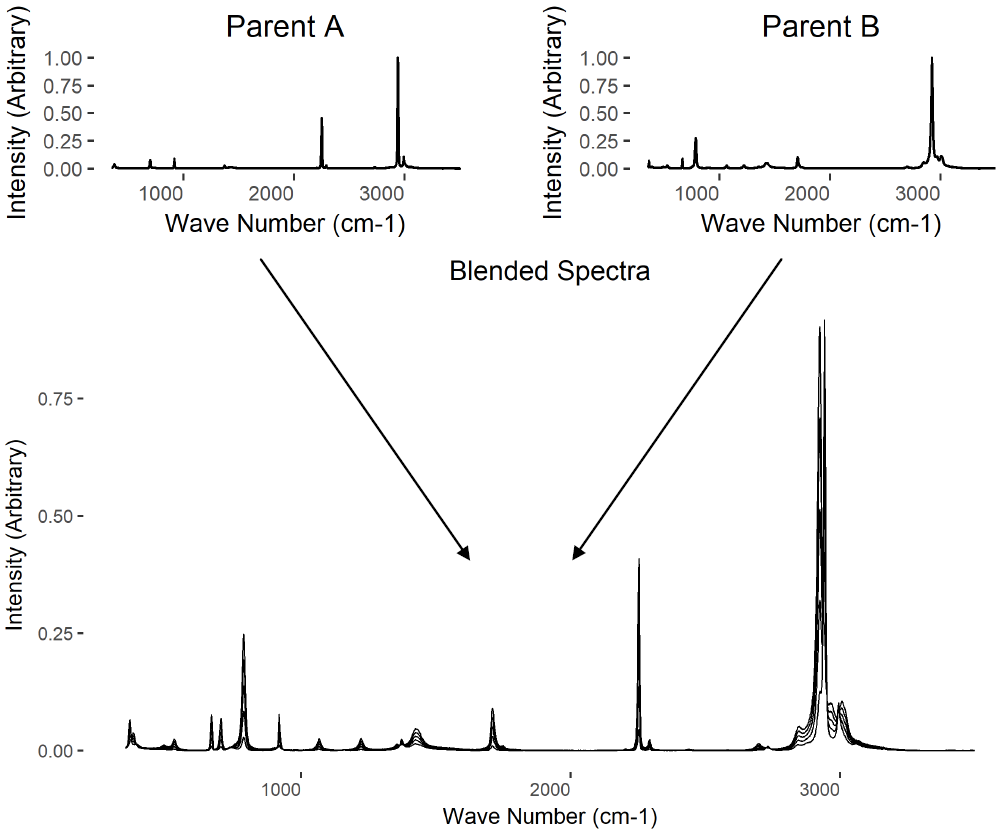}
\caption{Example result of blending two chlorinated spectra.}
\label{fig7}
\end{figure}

A spectrum blender method is used to produce label preserving synthetic data. Each of the five training folds are used to separately produce five sets of synthetic blended spectra. In this way the corresponding test folds can be used without any issues of cross contamination, (i.e. the same data used for generating the synthetic spectra being present in the test set). Since the synthesis process is label preserving in nature, each training fold is first separated by class and the positive and negative synthetic spectra are generated separately from the positive and negative training sets respectively. \\
\indent Generating a blended spectrum involves computing the weighted mean of the intensity of two or more parent spectra at each point in the spectrum. In this work, only binary combinations (two parent spectra per blended spectra) are considered. For each class of each training fold, every unique binary combination of training spectra is considered. For each binary combination, five blended spectra are produced by using different weightings of the parent spectra $A$ and $B$ as follows: 
\begin{itemize}
	\item $0.1 A + 0.9 B$  
	\item $0.3 A + 0.7 B$  
	\item $0.5 A + 0.5 B$  
	\item $0.7 A + 0.3 B$ 
	\item $0.9 A + 0.1 B$
	\end{itemize}
Figure \ref{fig7} illustrates the blending process for a single pair of parent spectra. The five resultant blended spectra appear to resemble the real spectra in structure, yet they are reasonably distinct from the exact form of either parent. 

Each training fold contains approximately 92 positive instances and 45 negative instances. The number of unique binary ordered permutations of a set of $n$ items is given by $\frac{1}{2} {}_n P_2 = n(n-1)/2$. Since five spectra are created per combination the number of blended spectra produced from a set of $n$ parent spectra becomes $5n(n-1)/2$. The approximate number of positive and negative blended spectra producible per training fold is shown below in Table \ref{table4}.

\begin{table}[h!]
\begin{adjustbox}{max width=0.46\textwidth}
\begin{tabular}{ | c | c | c | }
\hline
	 & Parent Spectra & Potential Blended Spectra \\ \hline
	Positive Class & $\approx$92 & $\approx$41,860 \\ \hline
	Negative Class & $\approx$45 & $\approx$9,900 \\ \hline
	Total & $\approx$137 & $\approx$51,760 \\ \hline

 \end{tabular}
 \end{adjustbox}
\caption{Number of possible blended spectra per fold}
\label{table4}
\end{table}

\begin{figure}[!h]
\centering
\includegraphics[width=3.3in]{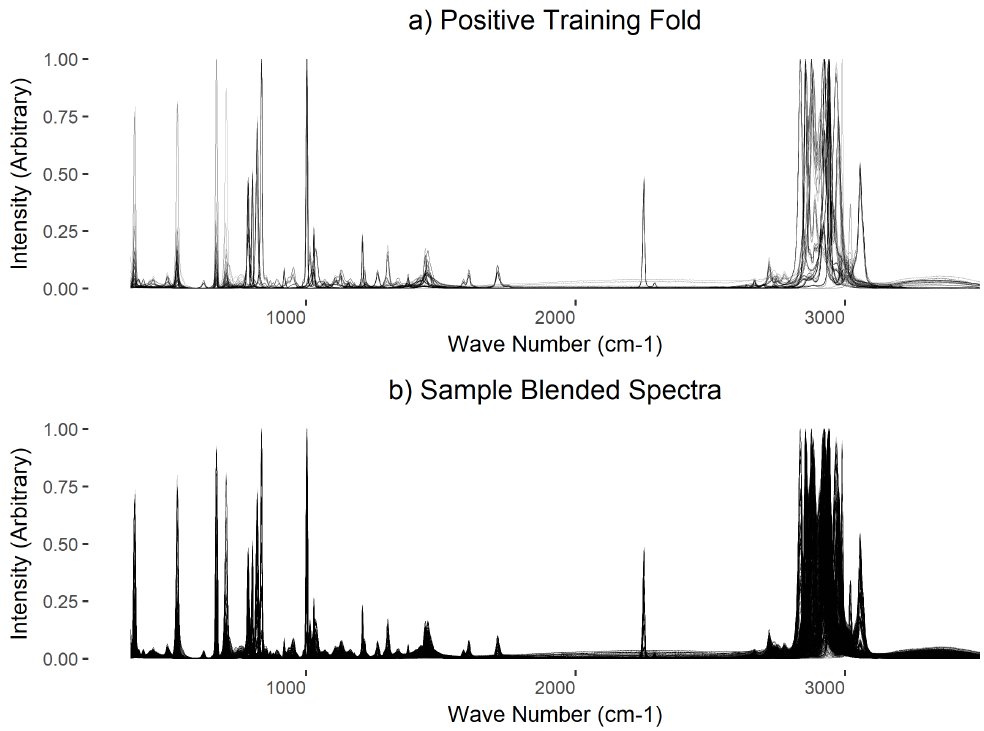}
\caption{Sample positive blended spectra produced from a single training fold.}
\label{fig8}
\end{figure}

Five sets of blended spectra are produced, corresponding to the five existing training folds. 
Figure \ref{fig8} shows: a) the positive spectra from a single training fold, and b) 2,000 randomly selected blended spectra which are generated from these positive spectra. 
  
Figure \ref{fig8} clearly shows the manner in which the blending process produces synthetic spectra which occupy the same subspace as the parent spectra, while increasing the density of spectra occupying this subspace. 
\subsubsection{Variational Autoencoder Data Generation}
\label{sec:vaeData}
\begin{figure}[!h]
\centering
\includegraphics[width=3.3in]{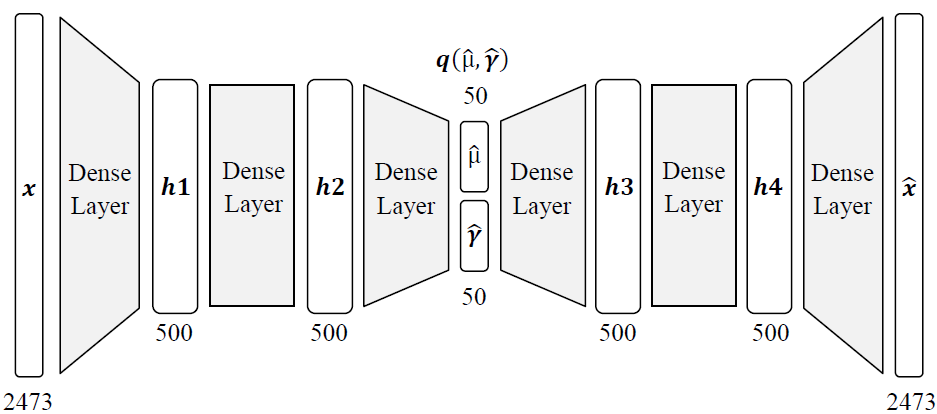}
\caption{Variational Autoencoder architecture.}
\label{fig9}
\end{figure}

A total of ten VAEs are trained and used for the generation of synthetic spectra, one per class label per training fold. The VAEs are constructed and trained using the TensorFlow package \cite{cite46}. The implementation is partially based on that found in Geron \cite{cite47}. All models use the same architecture shown in Figure \ref{fig9}.

\begin{figure}[!h]
\centering
\includegraphics[width=3.3in]{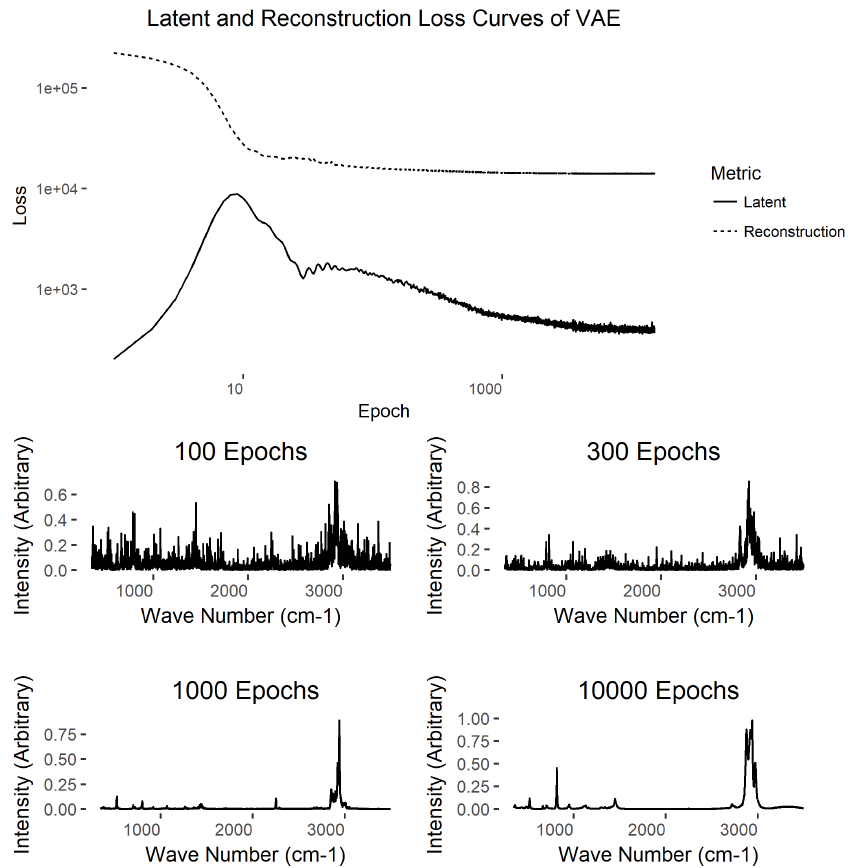}
\caption{a) Reconstruction and latent loss training curves of example VAE and b) sample generated spectra after different amounts of training.}
\label{fig10}
\end{figure}

The hidden layers h1, h2, h3, and h4 all use an Exponential Linear Unit (ELU) activation function\hl{, as other research has indicated its suitability} \cite{cite54}. The output layer uses sigmoid activation. The central mean and gamma layers have no activations. Gamma is used in place of sigma to better capture sigmas of different scale, $\gamma = log(⁡\sigma^2)$. The training cost function has two parts, a reconstruction loss and a latent loss. The reconstruction loss of the output is calculated based on the cross-entropy. 
The VAEs are trained using an Adam optimizer, with a learning rate of 0.001 for 10,000 epochs. 
Figure \ref{fig10} shows the reconstruction and latent loss curves used during training of an example VAE. Additionally, some sample generated spectra are shown after various numbers of epochs. It can be clearly seen that the synthetic spectra begin to more closely resemble real spectra over the course of training.

\begin{figure}[!h]
\centering
\includegraphics[width=3.3in]{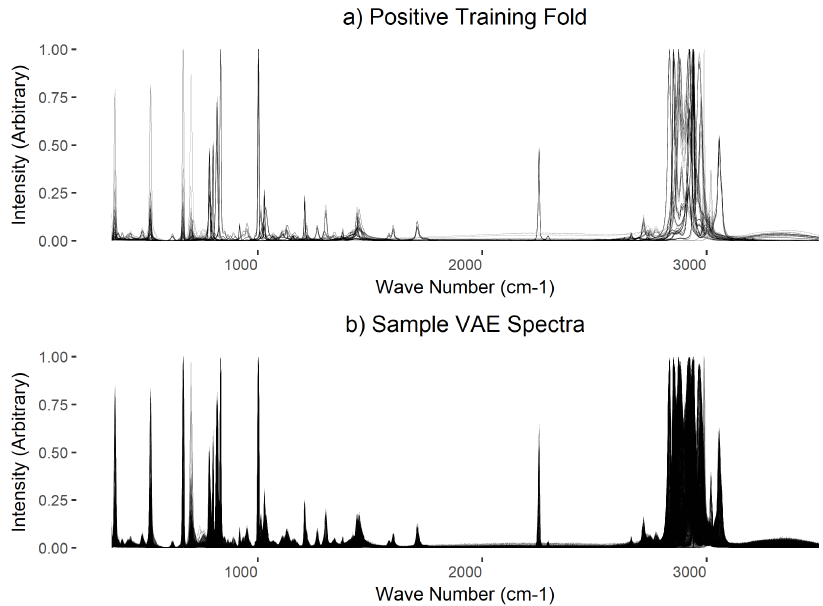}
\caption{Sample positive VAE spectra produced from a single training fold.}
\label{fig11}
\end{figure}

Figure \ref{fig11} shows a) the positive spectra from a single training fold, and b) 2,000 randomly selected spectra generated from a VAE model trained on these spectra. As with the blended spectra, the VAE generated spectra appear to more fully occupy the same subspace as the real spectra.
  
Five sets of VAE generated spectra are produced, corresponding the five existing training folds. Each set contains 10,000 synthetic spectra. 
\subsubsection{Testing Synthetic Data}
\label{sec:testSynthData}
A set of tests are performed to explore the effect on classification accuracy of including synthetic data in the training. The kNN, SVM, and LCNN models discussed earlier were used in the comparative analysis. The optimal parameters determined for each model and the architecture of the LCNN discussed earlier are used.
Each model is trained under five training data scenarios as follows: 
	\begin{itemize}
	\item Real spectra only 
	\item Blended spectra only
	\item Blended + real spectra
	\item VAE generated spectra only
	\item VAE generated spectra + real spectra.
	\end{itemize}
Additionally, where applicable, the number of synthetic spectra introduced into the training set is varied in a range between 30 and 10,000. The proportion of positive to negative instances of the added synthetic data is kept consistent with the original class balance found in the real data. Training and testing for each model and scenario is performed for each of the five training / testing folds with the mean and corresponding standard error of model accuracy on the test set being recorded. 
\subsection{Results}
\label{sec:results}
\subsubsection{Spectrum Blender Data}
\begin{figure}[!h]
\centering
\includegraphics[width=3.3in]{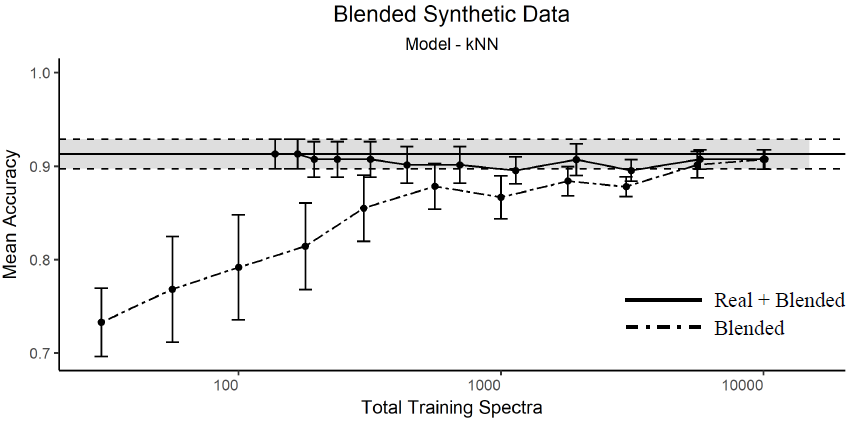}
\caption{Effect of using blended synthetic training data on kNN model accuracy.}
\label{fig12}
\end{figure}

\textbf{\\kNN:\\}
The effect of using blended synthetic training data performance of kNN models is tested. The same model parameters that yielded optimum performance are used here, ($k = 1$, distance metric = Manhattan). Three cases are considered, training on real data only, training on blended spectra only, and training on real and blended spectra combined.\\ 
\indent Figure \ref{fig12} shows a plot of the mean accuracies attained, along with associated standard errors. The best accuracy achieved for each scenario is summarized in Table \ref{table5}. The horizontal line and shaded region show the performance obtained by training on the real data only. The solid line series shows the model performance when blended synthetic training data is added to the real training data. No significant effect on performance is seen. The dashed line series shows the model performance when trained on the blended synthetic data only. At high quantities, training on just the synthetic data achieves comparable performance to training on real data. This indicates that the blended synthetic data has been produced in a label preserving manner, meaning that the positive and negative blended spectra contain the information necessary to distinguish unseen cases with reasonable accuracy. 

\begin{table}[h!]
\begin{adjustbox}{max width=0.46\textwidth}
\begin{tabular}{ | c | c | c | c | }
\hline
	\textbf{Data} & \textbf{Training Spectra} & \textbf{Mean Accuracy} & \textbf{Std. Error} \\ \hline
	Real & 138 & 0.913 & 0.016 \\ \hline
	Blended & 10000 & 0.907 & 0.010 \\ \hline
	Real + Blended & 10138 & 0.907 & 0.010 \\ \hline
\end{tabular}
\end{adjustbox}
\caption{kNN model testing with blended data}
\label{table5}
\end{table}

\begin{figure}[!h]
\centering
\includegraphics[width=3.3in]{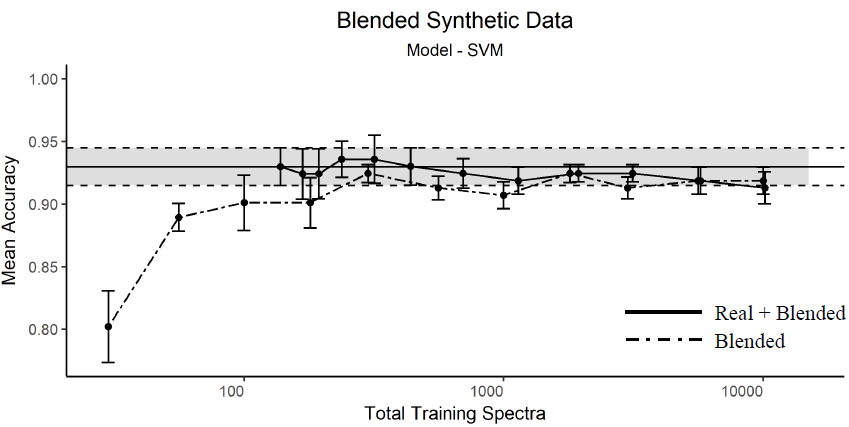}
\caption{Effect of using blended synthetic training data on SVM model accuracy.}
\label{fig13}
\end{figure}

\textbf{\\SVM:\\}
The effect of using blended synthetic training data on the performance of SVM models is tested. The same model parameters which yielded optimum performance are used here, ($C = 10$, kernel = Linear). Three cases are considered, training on real data only, training on blended spectra only, and training on real and blended spectra combined. \\
\indent Figure \ref{fig13} shows the plot of the mean accuracies attained, along with associated standard errors. The best accuracy achieved for each scenario is summarized in Table \ref{table6}. As with the kNN algorithm, the addition of blended synthetic data does not appear to have a significant effect on model performance. This is an unsurprising result considering the properties of an SVM. Good generalization is already promoted through finding of the maximal margin hyperplane. The addition of synthetic data is unlikely to significantly affect the location of the separating hyperplane if the positive and negative synthetic data come from similar distributions to the real data. The dashed line series showing model performance with just synthetic blended data shows near real data performance with only approximately 300 synthetic spectra. Again, this indicates that the blended synthetic data does contain the information necessary to learn a model which can make accurate predictions on real test data.
 
\begin{table}[h!]
\begin{adjustbox}{max width=0.46\textwidth}
\begin{tabular}{ | c | c | c | c | }
\hline
	\textbf{Data} & \textbf{Training Spectra} & \textbf{Mean Accuracy} & \textbf{Std. Error} \\ \hline
	Real & 138 & 0.93 & 0.015 \\ \hline
	Blended & 300 & 0.924 & 0.007 \\ \hline
	Real + Blended & 238 & 0.936 & 0.015 \\ \hline
\end{tabular}
\end{adjustbox}
\caption{SVM model testing with blended data}
\label{table6}
\end{table}

\begin{figure}[!h]
\centering
\includegraphics[width=3.3in]{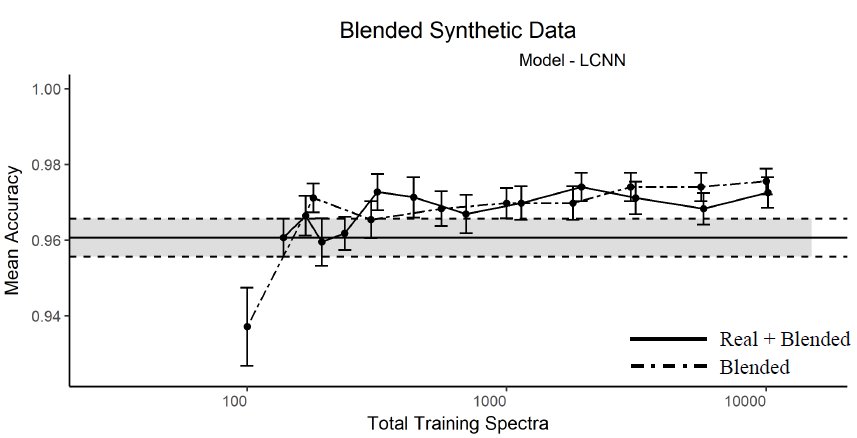}
\caption{Effect of using blended synthetic training data on LCNN model accuracy.}
\label{fig14}
\end{figure}

\textbf{\\LCNN:\\}
The effect of using blended synthetic training data on the performance of a LCNN model is tested. The same architecture and training parameters as described earlier are used here. Three cases were considered, training on real data only, training on blended spectra only, and training on real and blended spectra combined. \\
\indent Figure \ref{fig14} shows a plot of the mean accuracies attained, along with associated standard errors. The best accuracy achieved for each scenario is summarized in Table \ref{table7}. A small increase in accuracy can be seen at large numbers of total training spectra, both when using real data plus the blended data and when using the blended data only. We discussed earlier how the ratio of trainable parameters to training instances had a potential effect on increasing generalization and hence performance of the model. At the larger values of total training spectra which were tested, the number of training instances exceeds the number of parameters in the LCNN model (2,541). It is at these larger values that a consistent better-than-real training data accuracy is found. Comparison of the blended and real-plus-blended series in Figure \ref{fig14} also indicates that the total number of spectra used is a better indicator of model performance than whether real data is used in training or not.  
 
\begin{table}[h!]
\begin{adjustbox}{max width=0.46\textwidth}
\begin{tabular}{ | c | c | c | c | }
\hline
	\textbf{Data} & \textbf{Training Spectra} & \textbf{Mean Accuracy} & \textbf{Std. Error} \\ \hline
	Real & 138 & 0.961 & 0.005 \\ \hline
	Blended & 10000 & 0.976 & 0.003 \\ \hline
	Real + Blended & 10138 & 0.973 & 0.004 \\ \hline
\end{tabular}
\end{adjustbox}
\caption{LCNN model testing with blended data}
\label{table7}
\end{table}

\subsubsection{Variational Autoencoder Data}
\begin{figure}[!h]
\centering
\includegraphics[width=3.3in]{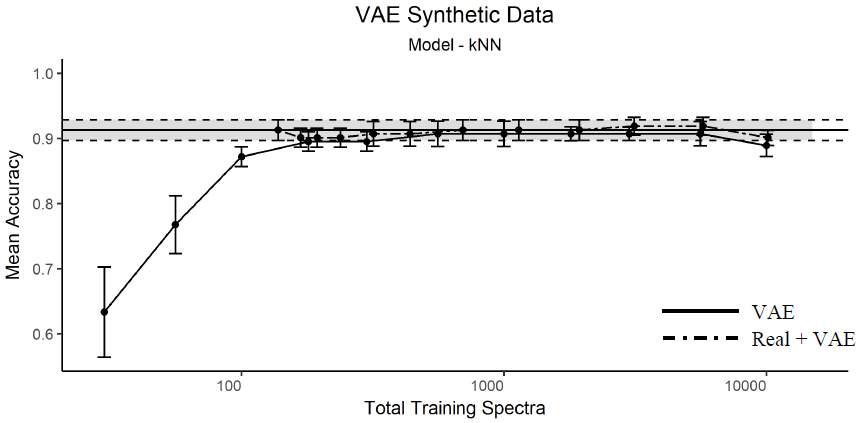}
\caption{Effect of using VAE synthetic data on KNN model accuracy.}
\label{fig15}
\end{figure}

\textbf{\\kNN:\\}
The effect of using VAE synthetic training data on the performance of kNN models is tested. The same model parameters that yielded optimum performance were used here, (k = 1, distance metric = Manhattan). Three cases were considered, training on real data only, training on VAE spectra only, and training on real and VAE spectra combined. \\
\indent Figure \ref{fig15} shows a plot of model accuracy against total number of training spectra. The best accuracy achieved for each scenario is summarized in Table \ref{table8}. As was the case with the blended data, adding VAE generated synthetic data to the real training data does not appear to have any significant effect on the model performance. The model performance when trained on only the synthesized data converges with real-training-data performance once the number of synthetic spectra approximately matches the number of real spectra. In comparison, when kNN models were trained on blended synthetic data only they did not reach parity with real-training-data performance until very large numbers of synthetic spectra were used, (5,000-10,000). Each VAE generated synthetic spectrum is in a sense a product of all of the spectra which were used to train the VAE. Thus information from all 138 training spectra is present in a set of, for example, 100 VAE generated spectra. Each blended spectrum however is a product of only two real spectra. For a random set of 100 blended spectra only approximately 77\% of the 138 real spectra will be represented by the synthetic data. This disparity in representation of the training data at lower numbers of synthetic spectra could explain the observed difference in behavior. 
 
\begin{table}[h!]
\begin{adjustbox}{max width=0.46\textwidth}
\begin{tabular}{ | c | c | c | c | }
\hline
	\textbf{Data} & \textbf{Training Spectra} & \textbf{Mean Accuracy} & \textbf{Std. Error} \\ \hline
	Real & 138 & 0.913 & 0.016 \\ \hline
	Blended & 10000 & 0.89 & 0.017 \\ \hline
	Real + Blended & 10138 & 0.901 & 0.012 \\ \hline
\end{tabular}
\end{adjustbox}
\caption{KNN model testing with VAE data}
\label{table8}
\end{table}

\begin{figure}[!h]
\centering
\includegraphics[width=3.3in]{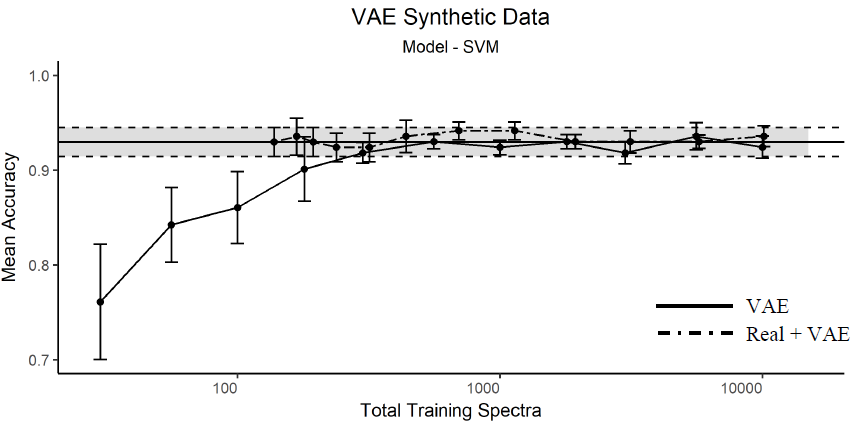}
\caption{Effect of using VAE synthetic training data on SVM model accuracy.}
\label{fig16}
\end{figure}

\textbf{\\SVM:\\}
The effect of using VAE synthetic training data performance of SVM models was tested. The same model parameters which yielded optimum performance were used here, (C = 10, kernel = Linear). Three cases were considered, training on real data only, training on VAE spectra only, and training on real and VAE spectra combined. 
Figure \ref{fig16} shows the plot of the mean accuracies attained, along with associated standard errors. The best accuracy achieved for each scenario is summarized in Table \ref{table9}. The use of VAE generated synthetic data does not appear to have any significant effect on SVM model accuracy. This is the same result as was observed when using blended data and is to be expected given the good generalization afforded by maximizing the separating hyperplane margin. 
 
\begin{table}[h!]
\begin{adjustbox}{max width=0.46\textwidth}
\begin{tabular}{ | c | c | c | c | }
\hline
	\textbf{Data} & \textbf{Training Spectra} & \textbf{Mean Accuracy} & \textbf{Std. Error} \\ \hline
	Real & 138 & 0.93 & 0.015 \\ \hline
	Blended & 5600 & 0.936 & 0.014 \\ \hline
	Real + Blended & 698 & 0.942 & 0.009 \\ \hline
\end{tabular}
\end{adjustbox}
\caption{SVM model testing with VAE data}
\label{table9}
\end{table}

\begin{figure}[!h]
\centering
\includegraphics[width=3.3in]{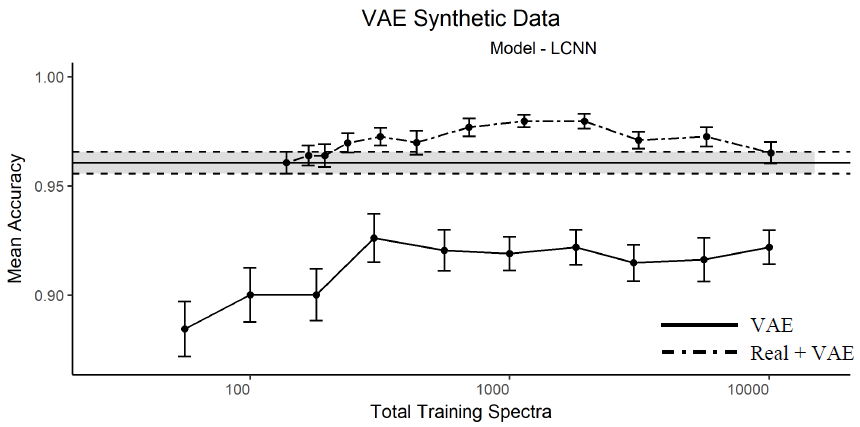}
\caption{Effect of using VAE synthetic training data on LCNN model accuracy.}
\label{fig17}
\end{figure}

\textbf{\\LCNN:\\}
The effect of using VAE synthetic training data performance of LCNN models is tested. The same model architecture described earlier is used. Three cases are considered, training on real data only, training on VAE spectra only, and training on real and VAE spectra combined. \\
\indent Figure \ref{fig17} shows a plot of the mean accuracies attained, along with associated standard errors. The best accuracy achieved for each scenario is summarized in Table \ref{table10}. Training on VAE generated synthetic data does not appear to be sufficient in producing real-data-training performance with the LCNN model, even at large numbers of total training spectra. This is in contrast to blended data which not only match real-data-training performance but exceeded it at high enough numbers of blended spectra. This would indicate that some pattern, important to the good performance achieved with LCNN models, which exists in the training data is not being captured by the VAE generated synthetic data. Despite this, adding VAE data to real training data does appear to improve classification accuracy within a certain range of added spectra, (approx. 560-3,000). At higher numbers of added spectra the performance begins to decrease again. This behaviour could be explained by the concurrence of two antagonistic trends. 1) Adding synthetic spectra reduces the ratio of trainable parameters to training instances, promoting better generalization and performance. 2) Adding large numbers of synthetic data reduces the ratio of real spectra to VAE generated spectra. As it has been shown that training on VAE generated spectra does not give as good performance with this algorithm as training with real data, it is reasonable to expect that this would degrade the performance if the model. 
 
\begin{table}[h!]
\begin{adjustbox}{max width=0.46\textwidth}
\begin{tabular}{ | c | c | c | c | }
\hline
	\textbf{Data} & \textbf{Training Spectra} & \textbf{Mean Accuracy} & \textbf{Std. Error} \\ \hline
	Real & 138 & 0.961 & 0.005 \\ \hline
	Blended & 300 & 0.926 & 0.011 \\ \hline
	Real + Blended & 1138 & 0.98 & 0.003 \\ \hline
\end{tabular}
\end{adjustbox}
\caption{LCNN model testing with VAE data}
\label{table10}
\end{table}

\subsubsection{Determining the `Best' Training Data Source for each Model}
From the results presented in the previous two sections, it can be seen that for the kNN and SVM algorithms the use of neither blended nor VAE generated synthetic data gives any improvement in model accuracy over training on real data only. Therefore there is no observed justification in using synthetic data and for all subsequent use of kNN and SVM models in this work only real data will be used in training. \\
\indent In the case of the LCNN algorithm, evidence of improved performance with the use of synthetic data is found. The highest overall mean accuracy is attained when training on a combination of real data and 1,000 VAE generated (mean accuracy 0.980$\pm$0.003). However this is within error of the highest mean accuracies achieved using blended spectra (0.976$\pm$0.003). Additionally, blended synthetic data show a much more robust effect of increasing model performance, particularly at higher total numbers of training spectra. For these reasons a combination of real data and 10,000 blended spectra is chosen as the training data source for all subsequent uses of LCNN in this paper.
\subsubsection{Summary of Results}
\label{sec:resultSummary}
The addition of synthetic training data has no observed effect of increasing the classification accuracy of either the kNN or SVM models. However the LCNN model, which already produces the best results using only real training data, does show improved performance with synthetic training data in certain circumstances. It is hypothesized that the addition of appropriate synthetic data reduces the tendency of the models to overfit to the limited training data, thus increasing the generalization and performance of the model. \\
\indent The most robust effect on classification accuracy is found with large volumes of blended spectra.

\section{Outlier Data and One-Class Classification}
\label{sec:occAnom}
\subsection{Introduction}
In this section, the presence of negative test instances which are not well characterized by the training data is considered. Such instances are here referred to as negative outliers. We begin by demonstrating the effect of negative outliers on the classification accuracy of the previously-presented binary classification models. \\
\indent As discussed earlier, one-class classification methods have been noted for their robustness to negative outliers. In this work, the suitability of autoencoder based OCC algorithms in discriminating chlorinated Raman spectra from all other Raman spectra is explored.\\
\indent Closely related to one-class classification is the task of outlier detection. In this work, outlier detection is considered as the task of discriminating between positive or negative instances from the principal dataset, and outlier negative instances. The same autoencoder algorithms which are tested as OCC algorithms are also employed and tested as outlier detectors. The relative performance of the same algorithms at these two distinct tasks is explored.
\subsubsection{Outlier Dataset}
An additional dataset comprising 24 Raman spectra of carbohydrates was compiled for use as examples of possible outlier data, i.e. non-chlorinated solvents containing samples which are \hl{\emph{unrepresented}} by the existing dataset. \hl{Spectra of samples of various monosaccharides, disaccharides, and polysaccharides were chosen at random from the SPECARB database} \cite{cite48}. \hl{These carbohydrates represent a class of substances that is quite distinct from the training dataset of solvents, but as the results below (Figure 21) will show, they cause significant problems for binary classifiers.}\\ 
\indent The obtained spectra contain Raman intensity readings at wavenumbers from 0-3600 cm-1 with a resolution of 1 cm-1. In order to allow for testing of our models with the outlier data, each spectra was interpolated to contain intensity readings at the same set of wavenumbers as the principal dataset. 
\begin{figure}[!h]
\centering
\includegraphics[width=3.3in]{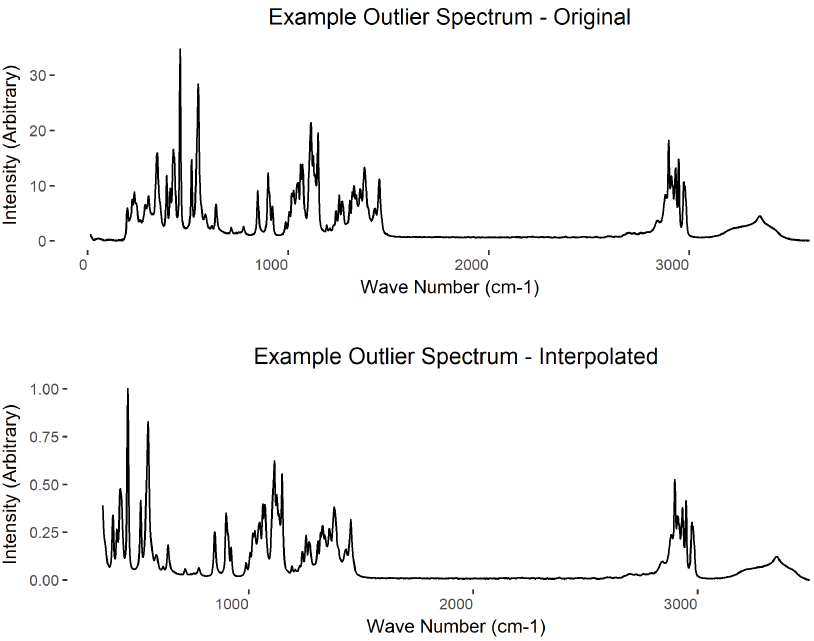}
\caption{Example outlier spectrum. The original (top) and the interpolated and truncated spectrum (bottom).}
\label{fig18}
\end{figure}
Figure \ref{fig18} shows an example spectrum from the outlier dataset. The form of the outlier is noticeably distinct from those of the principal dataset. Comparison of the original (top) and interpolated (bottom) spectra show that the interpolation process has accurately maintained the spectral form.
\subsubsection{One-Class Classification Algorithms}
One-Class classification (OCC), also known as one-sided or single-class classification is an alternative to traditional multi-class classification, as was discussed earlier. It differs in that the classifier is trained to recognise only a single target class, often referred to as the positive class. The classifier learns to distinguish instances of the target class from all other instances in the feature space. One-Class classifiers share similarities and applications with anomaly and outlier detection in that a single, well defined subset of instances is distinguished from all other data. \\
\indent A primary motivation for the use of OCCs is the difficulty of obtaining representative negative training data in many domains. In such cases, a binary classifier may perform poorly when presented with negative test instances that were not well represented in the training data. Consider the case here of determining if any of a particular set of solvents (chlorinated) is present in a sample based on its Raman Spectrum. A negative training set would need to be vast in order to be a good representation of all of the possible chemical mixtures not containing the target solvents. OCCs avoid this issue by only using positive training data. Two OCC algorithms are tested and compared in this work, one based on an autoencoder, and another on a denoising autoencoder.\\
\textbf{Autoencoder as a One-Class Classifier: }\\
Recent work has seen the development of novel OCC algorithms based on autoencoders \cite{cite49}. As previously discussed, an autoencoder is a feed-forward neural network which is trained to reproduce its input as output. The use of autoencoders as OCC algorithms is based on the principle that a restricted autoencoder cannot learn an identity function to reproduce arbitrary input data. It can only learn to effectively reproduce data from a subset of the possible input data space through exploiting underlying structures in the data. By training only on a subset of data (the target class) the ability of the autoencoder to reproduce an unseen instance can be used as a test of whether the instance belongs to the trained on class. The restriction on the autoencoder is typically achieved through restriction of the encoded dimensionality, i.e. a hidden layer with a small number of nodes compared to the input / output. Other types of restriction are also possible.
\textbf{Denoising Autoencoder as a One-Class Classifier: }\\
A Denoising autoencoder recreates the input from a corrupted (noisy) version of the input. The added step of removing noise prevents the autoencoder from learning a simple identity function to map input to output and can have the effect of forcing the network to learn more robust features. As with the traditional autoencoder, the ability of the denoising autoencoder to reproduce an unseen instance can be used to determine if an unseen instance belongs to the positive class (training class). \\
\subsubsection{Outlier Detection}
Outlier Detection, also known as anomaly detection, refers to the detection of instances which do not conform to the same pattern as the rest of the dataset. In a machine learning context, it often describes the process of detecting instances which do not belong to the same population subset from which the training data was derived (i.e. test instances that were not effectively characterized in the training phase).  It is closely related to one-class classification in the sense that a single, well defined subset of instances is distinguished from all other data. \\
\indent In this work, the OCC algorithms described above (Autoencoder based OCC, Denoising Autoencoder based OCC) are also employed as outlier detection algorithms by simply training the algorithms on both the positive and the negative training instances with all instances treated as belonging to the target class (non-outliers). 
\subsection{Methodology}
\subsubsection{Implementation of the OCC Algorithms}
\begin{figure}[!h]
\centering
\includegraphics[width=3.3in]{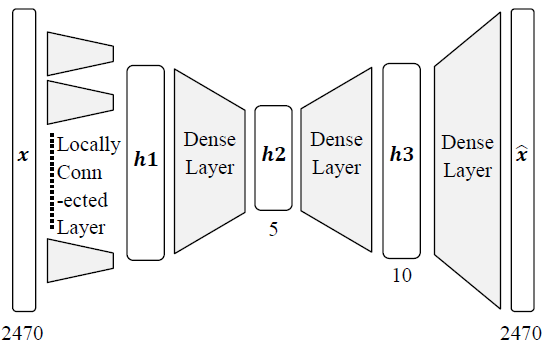}
\caption{Autoencoder architecture used in one-class classification algorithm.}
\label{fig19}
\end{figure}

\begin{figure}[!h]
\centering
\includegraphics[width=3.3in]{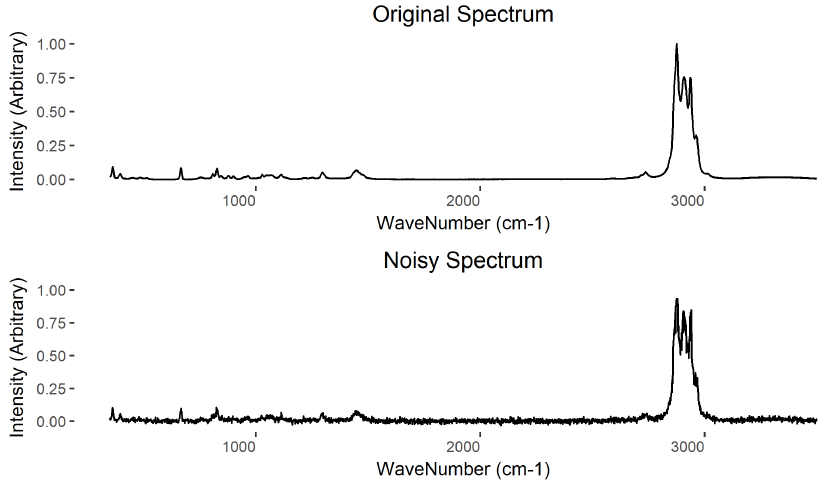}
\caption{Example of an original ``clean'' spectrum (above) and the same spectrum with noise added (below).}
\label{fig20}
\end{figure}

\hfill\\
\textbf{Autoencoder as a One-Class Classifier: }\\
The autoencoder architecture used is implemented using the keras package \cite{cite45}. Figure \ref{fig19} shows the details of the architecture used. This architecture is chosen due to its similarity to the LCNN architecture employed earlier, with the encoding portion being identical to the LCNN with the final output layer removed. The decoding portion then simply has fully connected layers fanning out to 10 and then 2470 dimensions. 
A model using this architecture is trained on only positive training data so as to minimize the reconstruction error of the input spectra. After the training of the model is completed, the maximum reconstruction error of a training instance is recorded and used as the error threshold. Unseen instances are then passed through the trained autoencoder and the resultant reconstruction error is compared to that recorded threshold. If the error exceeds the threshold then the instance is classified as negative. If it does not exceed the threshold, it is classified as positive. \\
\textbf{Denoising Autoencoder as a One-Class Classifier: }\\
An identical keras model as before was used. However, in both the training and testing phases noise was added to each intensity value of each input instance. Two sources of noise were added, multiplicative and additive. A vector of length equal to the number of intensity values per spectrum was generated containing of random values from a mean = 1, variance = 0.1 Gaussian distribution. The values in this vector were multiplied with corresponding intensity values in a spectrum to generate the effect of multiplicative noise. Another vector containing random values from a mean = 0, variance = 0.01 Gaussian distribution. Adding this vector introduced the additive noise. Figure \ref{fig20} provides a comparison of a clean spectrum and the noisy version of the same spectrum. A perceptible corruption of the spectrum is seen, however the underlying structure of the spectrum remains largely intact. Additionally, the use of both additive and multiplicative noise can be seen to cause corruption at both high and low intensity regions of the spectrum. The reconstruction error in both training and testing was calculated compared to the original (no added noise) spectrum. As before the maximal training instance reconstruction error is used as the classification threshold. 
\subsubsection{Implementation of the Outlier Detection Algorithms}
\label{sec:impOutDet}
As the same algorithms are employed for outlier detection as were employed for OCC the implementation is very simple. The same methods as discussed above for an autoencoder based OCC and denoising autoencoder based OCC are used. However, both positive and negative instances were used in training. For an unseen instance the output is interpreted as either `\emph{has been determined to be an outlier}' or `\emph{has not been determined to be an outlier}'.
\subsubsection{Testing the Effect of Outlier Data on Binary Classification Performance}
\label{sec:testOutData}
The performance of the previously discussed binary classification models (kNN, SVM, and LCNN) are retested under the scenario is which negative outlier data is present in the test dataset. For the kNN and SVM algorithms only real training data is used. For the LCNN algorithm real training data plus 10,000 blended synthetic spectra is used.
Model accuracy testing is performed under four test set scenarios: 
\begin{enumerate}
\item Original test set only
\item Original test set + 8 negative outliers
\item Original test set + 16 negative outliers
\item Original test set + 24 negative outliers
\end{enumerate}
These same four test set scenarios are used for all subsequent tests. 
\subsubsection{Testing OCC Algorithm Performance}
The same three training data source scenarios as discussed in the previous section are used. The autoencoder OCC and denoising-autoencoder OCC models are each trained on only the positive class training data. Training is performed both using real data only, and using real data plus 10,000 blended spectra. 
The models' accuracies in classifying positive and negative test instances are then tested under varying numbers of negative outlier spectra as described previously. \\
\subsubsection{Testing Outlier Detection Algorithm Performance}
The same process is repeated but with the algorithms being trained and tested as outlier detectors. They are trained on all of the training data, positive and negative. The model's accuracies are then tested (ability to correctly label outliers and non-outliers). 
\subsection{Results}
\subsubsection{Effect of Outlier Data on Binary Classification Performance}
\begin{figure}[!h]
\centering
\includegraphics[width=3.3in]{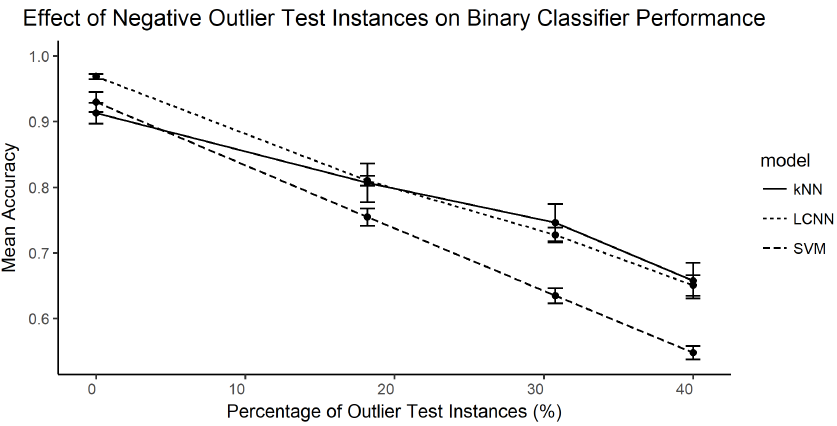}
\caption{Effect of Outlier Data on Binary Classification Accuracy.}
\label{fig21}
\end{figure}

The effects of introducing negative outliers to the test data on the mean accuracy of trained kNN, SVM, and LCNN models are shown in Figure \ref{fig21}. In all three cases the presence of outlier test instances is extremely detrimental to binary classification performance. Thus, although high classification accuracy was achieved with the LCNN model, this model is not robust to scenarios in which it is presented with new types of non-chlorinated samples. This shortcoming of binary classification models is consistent with the literature and is the primary motivating factor in the exploration of one-class classification and outlier detection techniques in relation to this dataset. 
\subsubsection{Performance of One-Class Classification Algorithms}
\begin{figure}[!h]
\centering
\includegraphics[width=3.3in]{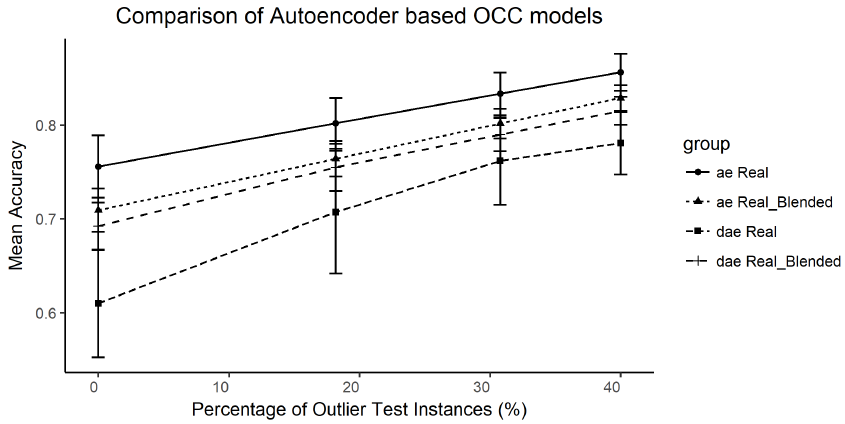}
\caption{Comparison of autoencoder based OCC models.}
\label{fig22}
\end{figure}

\begin{figure}[!h]
\centering
\includegraphics[width=3.3in]{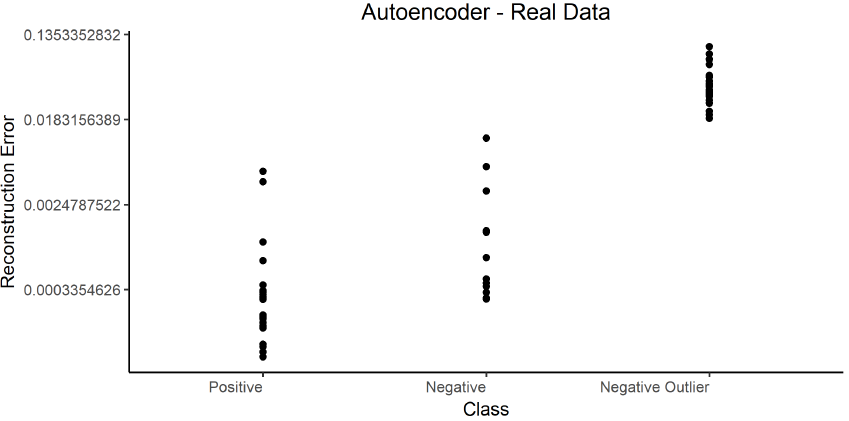}
\caption{Example Reconstruction errors of the Positive, Negative, and Negative Outlier instances in a test set.}
\label{fig23}
\end{figure}

Figure \ref{fig22} shows performance of the autoencoder and denoising-autoencoder based OCC models. In all cases, the performance on the original test set is substantially poorer than that achieved through binary classification. In contrast to the binary classifiers however, the addition of negative outliers to the test set improves the performance of the one-class classifiers. These results are consistent previous work on this dataset \cite{cite3}. The standard autoencoder, trained on real data yielded the highest average performance of the models tested, though other models performed within error of this. The classification accuracy attained through the use of autoencoder based OCC methods does not align with the classification accuracy reported by Glavin and Madden \cite{cite3} when using a standard one-sided kNN model. While these authors were using the same dataset, their experimental setup and classification algorithm was different to ours. To investigate the comparably poor performance of our approach, the reconstruction error of the test instances for an example autoencoder trained on real data is shown in Figure \ref{fig23}. The positive, negative, and negative outlier errors are shown separately. While the reconstruction error of negative instances is on average higher than that of positive instances, the large degree of overlap means that any choice of threshold will yield relatively poor classification accuracy. By contrast, the range of reconstruction errors found for negative outliers is quite distinct and can be well separated from both the positive and negative inliers. These observations explain both the poor overall performance of the autoencoder-based OCC models and the trend of improved performance with the addition of negative outliers to the test set. Additionally, they point to the potential suitability of these algorithms for outlier detection with respect to this data. 

\subsubsection{Performance of Outlier Detection Algorithms}
\begin{figure}[!h]
\centering
\includegraphics[width=3.3in]{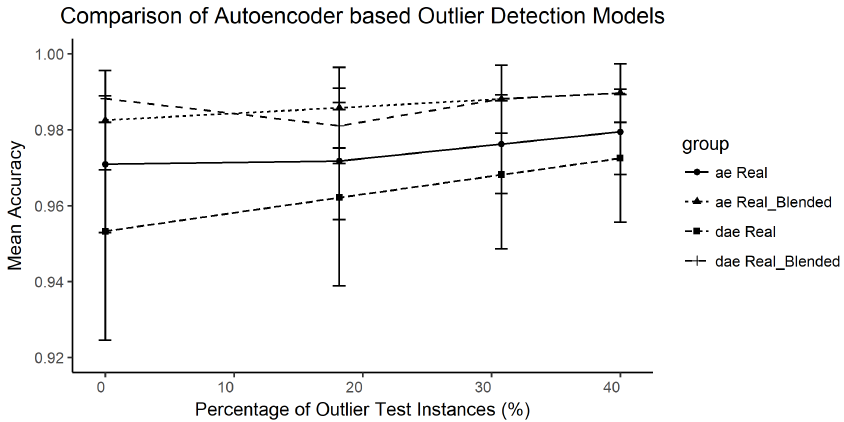}
\caption{Comparison of autoencoder based outlier detection models.}
\label{fig24}
\end{figure}

Figure \ref{fig24} shows performance of the autoencoder and denoising-autoencoder based outlier detection models. As expected the performance of the same algorithms at this task is greatly improved, compared to the more difficult task of one-class classification. No obvious improvement in performance is given by either the use of a denoising-autoencoder or with the use of synthetic data. 
\subsubsection{Summary of Results}
While the use of a LCNN architecture and the addition of large volumes of synthetic training data allows for accurate classification on test instances from the principal dataset, the introduction of negative outliers exposes a vulnerability of the binary classification model. Valid negative instances which have not been well characterized by the training set are often misclassified. Autoencoder based one-class classification algorithms show the reverse behavior. They accurately classify instances from the outlier dataset since they are sufficiently distinct from the positive class. However poor performance on the principal dataset is observed. The task of learning a subtle decision boundary when supported by data on one side only is more difficult than that of the binary classification paradigm. The much improved performance of the same autoencoder based algorithms when employed as outlier detectors suggests that this is an easier task to learn when only the target class is presented in training. 

\section{Two-Step Classification}
\label{sec:twoStep}
\subsection{Introduction}
\begin{figure}[!h]
\centering
\includegraphics[width=3.3in]{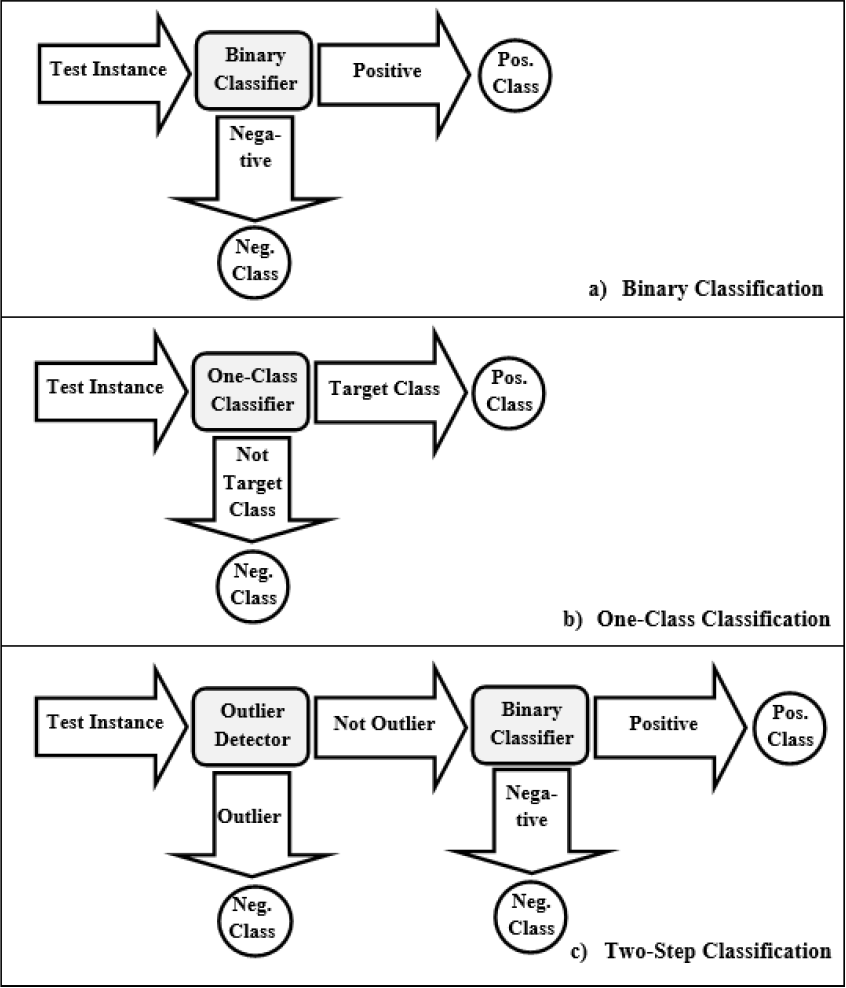}
\caption{Comparison of prediction processes in a) Binary Classification, b) One-Class Classification, and c) Two-Step Classification.}
\label{fig25}
\end{figure}

\begin{figure}[!h]
\centering
\includegraphics[width=3.3in]{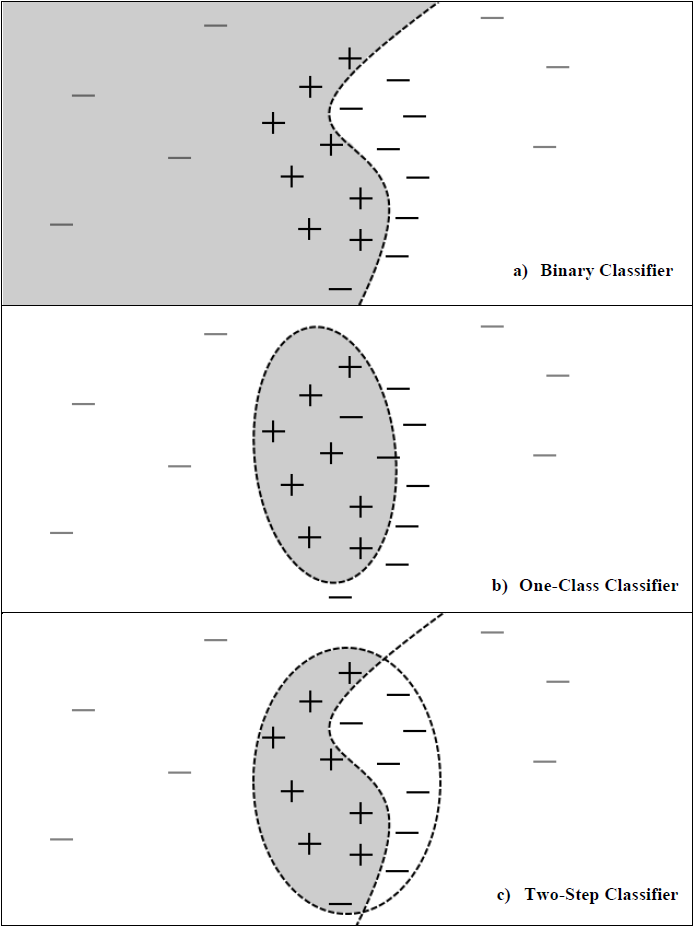}
\caption{Hypothetical illustration of the decision boundaries in a) Binary Classification, b) One-Class Classification, and c) Two-Step Classification.}
\label{fig26}
\end{figure}

Based on the results obtained in the previous section, we propose a two-step classification process as a method of achieving high classification accuracy which is also robust to the presence of large proportions of negative outlier instances. This two-step process is proposed as an alternative to a one-class classification procedure which may be employed under the same circumstances.\\
\indent The proposed model is motivated by three results which have been attained thus far:
\begin{enumerate}
\item High binary classification accuracy in the absence of outlier spectra
\item High outlier detection accuracy which is robust to large proportions of negative outlier data
\item Reduced accuracy of one-class classification techniques in predicting non-outlier data compared to binary classification models. 
\end{enumerate}
The two-step process is simple. Two models are trained: A binary classification model is trained on the labelled positive and negative training data. An outlier detection model is then trained on both the negative and positive training data (unlabeled). In the prediction phase an instance is first passed through the outlier detection model to detect negative outliers. In the case that the instance is not determined to be a negative outlier it is then passed on to the trained binary classifier. Thus the first step detects unrepresented negative instances (negative outliers), and the second step detects negative instances which were well represented by the training data, (non-outlier negatives). Figure \ref{fig25} provides a direct comparison of the classification processes followed in binary, one-class, and the proposed two-step classification methods. Whereas a one-class classifier attempts to determine all outliers in a single step, the two-step process makes use of available negative training data and decomposes the task into first finding negative outliers, and then finding negative instances represented by the training data. 
Figure \ref{fig26} provides a sketch illustrating of the decision boundary formed by a two-step classifier. The motivation is to combine the desirable properties of a one-class and binary classifier, while minimizing their drawbacks. That is, combine the robustness to negative outliers afforded by a OCC without sacrificing the good separation between well represented positive and negative instances obtainable through binary classification.

\subsection{Methodology}
\subsubsection{Implementation and Design Choices}
The two-step process comprises a binary classification algorithm and an outlier detection algorithm. The autoencoder based outlier detector discussed earlier is used. This is trained on real training spectra only. The LCNN is chosen for use in the binary classification step. The LCNN models are trained on real spectra plus 10,000 blended synthetic spectra. \\
\indent In the prediction phase, a test instance is first passed to the outlier detector. If the instance is determined to be an outlier then it is assigned to the negative class. If the test instance is not determined to be an outlier then it is passed on the binary classifier. The binary classifier then identifies it as belonging to the positive or negative class. 
\subsubsection{Testing}
The two-step classification process is tested on the five existing training and testing folds. Training and testing of each fold is repeated five times with different seeds for random number generation. Thus a total of 25 two-step models are trained and tested, with the mean accuracy and standard error of accuracy being recorded. Each model is tested under the four scenarios regarding the presence of negative outliers in the test set. \\
\indent In addition to this, a final model is trained using all of the non-validation set training data available. 10,000 blended synthetic spectra are compiled by taking 2,000 spectra from each of the 5 blended synthetic data folds created. This final model is then tested on the validation set of 59 spectra which has not been used in the training or testing of any models thus far. The 4 negative outlier scenarios are again used in testing. As before, each two-step model is trained and tested 5 times with different seeds for random number generation. 
\subsection{Results}
\subsubsection{Comparison between Binary and One-Class Classification}
\begin{figure}[!h]
\centering
\includegraphics[width=3.3in]{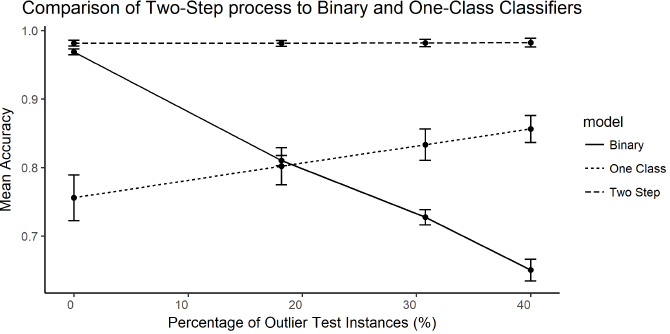}
\caption{Comparison of Two-Step classification process to Binary and One-Class alternatives.}
\label{fig27}
\end{figure}

Figure \ref{fig27} shows a plot of the classification accuracies achieved with the binary classification LCNN model, the one-class autoencoder model, and the two-step LCNN and autoencoder model. In contrast with the binary and one-class models, the two-step model shows excellent performance both with and without the presence of negative outliers. 
These results show that the two-step process is effectively combining the strengths of the OCC and binary classification alternatives. A particularly interesting result is the higher accuracy achieved with the two-step process over the binary classifier, even in the absence of any negative outlier test instances. Equal performance to the binary classifier in this scenario would indicate that the outlier detector is not producing any false negatives. The improved performance further indicates that certain negative test instances from the principal dataset, which are misclassified by the binary classifier, are being flagged as outliers and correctly classified.
\subsubsection{Testing the final Two-Step model on a Validation Set}
\begin{figure}[!h]
\centering
\includegraphics[width=3.3in]{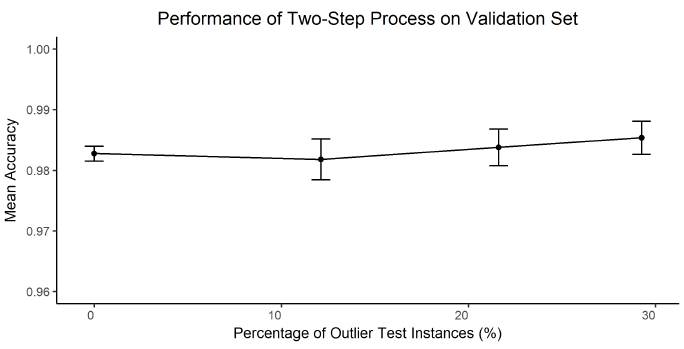}
\caption{Validated Two-Step model performance showing high accuracy which is robust to the presence of negative outliers.}
\label{fig28}
\end{figure}

A final test of the two-step process is performed by testing on a held-out validation portion of the principal dataset. Figure \ref{fig28} shows the mean accuracies achieved on the validation set both with and without the presence of negative outliers. Above 98\% classification accuracy is observed across the range of negative outlier test data percentages. These results provide validation of the LCNN binary classification model, the autoencoder outlier detector, and the two-step process by which they are combined. 

\section{Conclusion}
\label{sec:conclusions}
\subsection{Contributions}
The contributions made in this work are as follows:\\
\textbf{1)	A novel neural network application for classifying high-dimensional spectroscopic data.}\\
We proposed the use of locally-connected neural network architecture to perform a classification task on spectroscopic data. We illustrated the suitability of this architecture by noting the high classification accuracy achieved compared to SVM, kNN or fully-connected neural network models when using the raw input data. This result is achieved with a limited number of training instances, ($\approx$138 per model). 
This high accuracy is also achieved without the need for any additional pre-processing steps such as manual feature engineering or extraction. 
However, it is also noted that the LCNN model, as with the other binary classifiers tested, performs very poorly when negative outliers are present in the test data. \\
\textbf{2)	Techniques for generation of synthetic spectral training data to improve classification accuracy.}\\
The use of synthetic data to boost the performance of the LCNN models in a data-sparse domain has also been demonstrated. The data-hungry nature of neural networks and deep learning techniques often mean that the limiting factor in performance is the volume of high-quality training data that is available. In the \emph{Data Synthesis and the Effect on Classification Accuracy} section, we present evidence that synthetic data production techniques may be used to produce high volumes of additional training instances which boost the performance of the LCNN model. The same synthetic data were by comparison not shown to have any effect on the performance of the kNN and SVM models. The implications of this are two-fold:
\begin{itemize}
\item In data-sparse spectroscopic domains, the LCNN model and data synthesis may be used complimentarily to produce high accuracy classification models. 
\item The continued increase in LCNN performance at higher training data volumes implies it may be particularly suitable in spectroscopic domains where large quantities of training examples exist.
\end{itemize}
Of the two data synthesis techniques tested, the spectrum blender produced more consistent results with the LCNN model, (see Figure \ref{fig14} and Figure \ref{fig17}). In fact, no clear benefit to using the real data over blended data for training the LCNN model was observed. \\
\textbf{3)	Autoencoder-based one-class classification and anomaly detection classifiers that are robust to negative outliers.} \\
We observed poor performance of the autoencoder based one-class classification methods. While higher mean reconstruction error was observed for negative test instances over positive test instances, a well separating reconstruction error threshold did not exist. Better one-class classification performance has been previously reported on this dataset using other methods \cite{cite43}\cite{cite3}. However the general result of one-class classification yielding notably inferior results to binary classification when testing on well represented positive and negative data is a consistent observation in the literature. The ability of the autoencoder one-class classifier to correctly label negative outliers is demonstrated by the increased classification accuracy at higher percentages of negative outliers present in the test data. This is in contrast to binary classification models, which perform well in classifying well-represented positive and negative test data, but fail to consistently classify negative outliers correctly.\\
\indent The ability of the autoencoder models to correctly identify negative outliers is further demonstrated by testing their applicability as outlier detectors. Here the models are trained to recognize data which is well characterized by the positive and negative training data. The same algorithms achieve significantly higher accuracy in this task than in the one-class classification task. This indicates that for the datasets in question, from a target-class / non-target class machine learning perspective, the separation of negative outliers from all other data is an easier task than the separation of the positive and negative instances in the primary dataset.\\
\textbf{4)	A two-step classification technique for situations where negative training examples do not adequately represent the variety of negative instances that may exist.}\\
A two-step process, combining an outlier detector and binary classifier, has been proposed and tested on the available spectroscopic data. The two-step classifier has been constructed by combining an autoencoder outlier detector and a LCNN binary classifier. Figure \ref{fig27} shows that the two-step process outperforms both the LCNN binary classification and autoencoder one-class classification algorithms when used individually, regardless of what percentage of the test data comprises negative outliers. 
\subsection{Future Work}
Further testing of this model architecture on other spectroscopic datasets would demonstrate if this method can be of broad applicability in the spectroscopic and chemometric domains.\\
\indent Additionally, it is possible that more extensive optimization of the LCNN model architecture could yield improved results, particularly when combined with the use of large numbers of synthetic training spectra. For example, when using locally-connected layers it is common to use multiple output channels. In this work, only one output channel was used in the locally-connected layer. \\
\indent We tested two methods of producing synthetic spectral data. There is scope for further testing of other data synthesis methods or of variations of the methods used. In particular the architecture of the variational autoencoder used could possibly be optimized in order to generate more realistic synthetic data.\\
\indent The demonstrated two-step classification process is proposed as an alternative to one-class classification in domains where the existing negative training data does not represent the full breadth of negative instances which are likely to be encountered. Thus, the applicability of this process is not necessarily limited to the spectroscopy domain, or the particular binary classification and outlier detection algorithms that were used in this work. 

\bibliographystyle{achemso}
\bibliography{references}
\end{document}